# Observability of Strapdown INS Alignment: A Global Perspective

YUANXIN WU
HONGLIANG ZHANG
MEIPING WU
XIAOPING HU
DEWEN HU, Senior Member, IEEE
National University of Defense Technology
People's Republic of China

Alignment of the strapdown inertial navigation system (INS) has strong nonlinearity, even worse when maneuvers, e.g., tumbling techniques, are employed to improve the alignment. There is no general rule to attack the observability of a nonlinear system, so most previous works addressed the observability of the corresponding linearized system by implicitly assuming that the original nonlinear system and the linearized one have identical observability characteristics. Strapdown INS alignment is a nonlinear system that has its own characteristics. Using the inherent properties of strapdown INS, e.g., the attitude evolution on the SO(3) manifold, we start from the basic definition and develop a global and constructive approach to investigate the observability of strapdown INS static and tumbling alignment, highlighting the effects of the attitude maneuver on observability. We prove that strapdown INS alignment, considering the unknown constant sensor biases, will be completely observable if the strapdown INS is rotated successively about two different axes and will be nearly observable for finite known unobservable states (no more than two) if it is rotated about a single axis. Observability from a global perspective provides us with insights into and a clearer picture of the problem, shedding light on previous theoretical results on strapdown INS alignment that were not comprehensive or consistent.. The reporting of inconsistencies calls for a review of all linearization-based observability studies in the vast literature. Extensive simulations with constructed ideal observers and an extended Kalman filter are carried out, and the numerical results accord with the analysis. The conclusions can also assist in designing the optimal tumbling strategy and the appropriate state observer in practice to maximize the alignment performance.

Manuscript received December 25, 2008; revised March 25, 2010 and August 25, 2010; released for publication September 7, 2010.

IEEE Log No. T-AES/48/1/XXXXXX.

Refereeing of this contribution was handled by J. Morton.

This work was supported in part by the National Natural Science Foundation of China (60604011), the Foundation for the Author of National Excellent Doctoral Dissertation of People's Republic of China (FANEDD 200897), Program for New Century Excellent Talents in University (NCET), and the Cheung Kong Scholars Program.

Authors' address: Department of Automatic Control, College of Mechatronics and Automation, National University of Defense Technology, Changsha, Hunan, People's Republic of China 410073, E-mail: (yuanx_wu@hotmail.com).

## I. INTRODUCTION

The strapdown inertial navigation system (INS) necessitates an alignment stage to determine the initial condition before the navigation operation. Alignment is vitally important, because the performance of an INS is largely decided by the accuracy and rapidness of the alignment process. We care most about the initial body attitude during the alignment, because other initial conditions such as position and velocity are relatively easy to determine [1–3]. Attitude is essentially an SO(3) manifold[1] [4–6], which inevitably brings strong nonlinearity to the strapdown INS system, as well as the alignment stage, and even stronger nonlinearity when inertial sensor (gyroscope and accelerometer) biases are considered or when rotating or accelerating maneuvers are intentionally introduced to improve the observability or estimability [7–10].

Regardless of the technique (gyrocompassing, state estimation, or observer) used to address the alignment problem, observability analysis is necessary, because it reveals the inherent estimability of the system [11–13]. For an unobservable system, we cannot achieve satisfactory estimation even if the measurement is accurate enough. Unfortunately, no formal criterion tells whether or not a nonlinear dynamic system, such as alignment and other strapdown INS-related problems, is observable [13], so observability works so far have been largely devoted to observability of the corresponding linearized system (direct form) or the linear error dynamic equation (indirect form) [7–10, 14–25]. For a linear time-invariant system, the observability analysis is straightforward: it tests the rank of the observability matrix. In contrast, the linearization is generally an implicit time-varying linear system, analysis of which is cumbersome and involves the evaluation of the observability Grammian [12]. If a linear time-varying system could be well approximated by a piecewise linear constant system and certain conditions on the null space of the dynamic matrix were met for each constant segment, the observability analysis would be considerably simplified, obtaining the main observability characteristics of the original system by a rank test of concatenation of the constant segment observability matrices [20], e.g., in transfer or in-flight alignment [20, 23], so-called multiposition alignment [7, 25], and simultaneous localization and mapping [26]. A general linear time-varying model was used in [9] to investigate the observability properties of INS/GPS (Global Positioning System) by examining the time derivatives of the system output.

---

[1]In differential geometry, SO(3) is the abbreviation for the special orthogonal matrix in three-dimensional space that has +1 determinant. The manifold can be simply seen as a linear space satisfying nonlinear constraints; specifically, the SO(3) manifold is a $3 \times 3$ matrix that satisfies the orthogonal and unit-determinant constraints.

Regarding the observability analysis based on the linearized system, we need to underline three weak points:

1) Practitioners often find themselves lost in nontrivial symbolic matrix rank testing, especially for high-dimensional systems [8, 9, 25]. It is extremely difficult to obtain analytical observability conditions for general linear time-varying systems, and practitioners must seek nonanalytical support from numerical simulations [9].

2) Linearization implies that the observability result can only locally characterize the properties of the original nonlinear system [13, 27, 28]. That is to say, the insights thus obtained are for the corresponding linearized system, not for the nonlinear system. Just as different curves might have the same tangential line, different nonlinear systems might have an identical linearized system, so the observability result for the linearized system may not be comprehensive for the original nonlinear system [29].

3) Setting aside the linearization approximation, validity of the observability analysis for the linearized system is arguable. By definition, a system is observable if the initial state can be determined given the state transition and measurement models of the system and outputs during some time interval [12, pp. 153–158]. In other words, the state and measurement matrices should be known. In [7–10] and [14–25], however, state, measurement, or both matrices of the linearized system are functions of the unknown current state at which the nonlinear system is linearized. It is the unknown state that is to be determined by the estimation process.

In this paper, we revisit the observability of strapdown INS alignment from a global perspective in an effort to overcome the preceding weaknesses. In particular, the statement that no general rule exists to check the observability of a general nonlinear system does not imply the impossibility of observability analysis by exploiting the structure inherent in special classes of systems [30]. Strapdown INS alignment, as well as other strapdown INS-based systems, is a kind of special-structure system, with its attitude state evolving on the SO(3) manifold. In earlier studies, we used the special SO(3) structure to examine the observability of nonlinear INS and odometer self-calibration [31, 32] and the INS/GPS system [29], yielding new, comprehensive insights. In this paper, we investigate the observability property of the original nonlinear strapdown INS alignment directly, starting from the basic definition of observability. Throughout this paper, we use the terms "global observability" to name the observability analysis of the nonlinear system on a finite time interval and "instantaneous observability" to name the observability analysis in an infinitely small neighborhood at the linearization point.[2] We know that instantaneous observability deals with the ability to distinguish the states from their neighbors in an infinitely small time interval or instantaneously, while global observability describes the ability to estimate the states in the entire time span [28]. The instantaneous observability concept is identical to that of global observability for a linear system but different for a nonlinear system. An instantaneously observable system is globally observable, but a globally observable system may be locally or instantaneously unobservable [28]. In other words, the requirements for global observability are looser than those for local ones. As we demonstrate, the global observability perspective is straightforward and constructive, leading to insights into and comprehensive understanding of tumbling effects on the nonlinear strapdown INS alignment problem.

This paper is organized as follows. Section II formulates and establishes strapdown INS alignment as a problem of solving a set of infinite nonlinear equations on a continuous time interval, in contrast to that of matrix rank computation in linearization-based observability analysis. Section III presents the result of global observability in the form of constructive theorems. Static and attitude-maneuvering cases are both considered. Sufficient conditions to make the alignment fully observable are analytically derived, drawing a clear picture of the effect of attitude maneuvers, i.e., inputs, on state observability. Section IV carries out extensive numerical simulations to aid understanding of the theoretical analysis. Simulation results accord with what theorems tell us. Conclusions are made in Section V.

The contribution of the paper is twofold. First, a global-observability perspective is proposed to investigate strapdown INS alignment, which provides us with insights into the problem and unveils the incompleteness and inconsistency of previous linearization-based observability studies of the problem. Specifically, the reporting of inconsistency calls for a review of all linearization-based observability studies in the vast literature. Second, this paper, along with [29], [31], and [32], provides a straightforward and efficient way to perform observability analysis for other strapdown INS-based systems.

## II. PROBLEM STATEMENT OF ALIGNMENT AND OBSERVABILITY

This section presents a mathematical formulation of strapdown INS alignment based on which observability analysis is to be performed in the sequel. Here, we focus on the ground alignment at a known location (longitude, latitude, and height are given). For brevity, the nonlinear alignment system is directly provided and the development details are readily available in textbooks such as [1–3] and [33].

---

[2]The previous linearization-based observability analysis is an approximation of instantaneous observability, but we show that it is not a consistent approximation.

Without loss of generality, the local-level frame $N$ is selected as the reference frame (east, north, up). We denote with $B$ the INS body frame, with $E$ the Earth frame, and with $I$ some chosen inertial frame. The inertial sensor outputs are contaminated by random constant biases. Using gyroscopes and accelerometers outputs, the ground velocity $\mathbf{v}^n = [v_N \; v_U \; v_E]^T$ and the body attitude matrix with respect to the reference frame $\mathbf{C}_n^b$ satisfy the kinematic equations as[3]

$$\dot{\mathbf{C}}_b^n = \mathbf{C}_b^n(\omega_{nb}^b \times) \quad (1)$$
$$\omega_{nb}^b = \omega_{ib}^b - \mathbf{b}_g - \mathbf{C}_n^b(\omega_{ie}^n + \omega_{en}^n)$$

and

$$\dot{\mathbf{v}}^n = \mathbf{C}_b^n(\mathbf{f}^b - \mathbf{b}_a) - (2\omega_{ie}^n + \omega_{en}^n) \times \mathbf{v}^n + \mathbf{g}^n \quad (2)$$

where $\omega_{nb}^b$ is the body angular rate with respect to the reference frame, expressed in the body frame; $\omega_{ib}^b$ is the error-contaminated body angular rate measured by gyroscopes in the body frame; $\omega_{ie}^n = [\Omega \cos L \; \Omega \sin L \; 0]^T$ is the Earth rotation rate in the reference frame, with $\Omega$ being the Earth rate and $L$ being the local latitude; $\omega_{en}^n = [v_E/(R_E + h) \; v_E \tan L/(R_E + h) \; -v_N/(R_N + h)]^T$ is the angular rate of the reference frame with respect to the Earth frame, expressed in the reference frame; $R_E$ and $R_N$ are, respectively, the transverse radius and the meridian radius of curvature; $h$ is the altitude; $\mathbf{f}^b$ is the error-contaminated specific force measured by accelerometers in the body frame; $\mathbf{g}^n = [0 \; -g \; 0]^T$ is the gravity vector in the reference frame; and $g$ is the magnitude of local gravity. The $3 \times 3$ skew symmetrical matrix $(\cdot \times)$ is defined so that the cross product satisfies $\mathbf{a} \times \mathbf{b} = (\mathbf{a} \times)\mathbf{b}$ for arbitrary two vectors. The gyroscope drift $\mathbf{b}_g$ and the accelerometer bias $\mathbf{b}_a$ are taken into consideration approximately as random constant vectors, i.e.,

$$\dot{\mathbf{b}}_g = \mathbf{0}, \quad \dot{\mathbf{b}}_a = \mathbf{0}. \quad (3)$$

Equations (1)–(3) form the augmented system dynamic equation, with its state comprising the body attitude matrix $\mathbf{C}_n^b$, the ground velocity $\mathbf{v}^n$, the gyroscope drift $\mathbf{b}_g$, and the accelerometer bias $\mathbf{b}_a$. Because the strapdown INS has zero ground velocity during ground alignment (whether angular motion exists or not), the measurement equation is

$$\mathbf{y} = \mathbf{v}^n \equiv \mathbf{0}. \quad (4)$$

The purpose of alignment is to estimate the state of system (1)–(4) using some kind of observer or estimation method. To attain accurate alignment, the observability analysis is indispensable, because it fundamentally reveals how to enhance the potential alignment performance through tumbling techniques.

This paper considers deterministic observability [11–13], where random noises in system dynamics and measurements are not considered, in contrast to stochastic observability evaluated under uncertainty [34–36]. A system is said to be observable if the initial state could be derived from knowledge of the system in finite time. In a more formal language, the definition of observability is as follows [12]:

"A system is said to be (globally) observable if for any unknown initial state $\mathbf{x}(0)$ there exists a finite $t_1 > 0$ such that knowledge of the input and the output over $[0, t_1]$ suffices to determine uniquely the initial state $\mathbf{x}(0)$. Otherwise, the system is said to be (globally) unobservable."

The ground velocity is observable, because $\mathbf{v}^n(0) = \mathbf{0}$. By substituting (4), the strapdown INS alignment observability problem (SAOP) of interest is reduced to the following:

Does it suffice to uniquely determine the initial state by solving the infinite nonlinear equations over the continuous time interval $[0, t_1]$

$$\dot{\mathbf{C}}_b^n = \mathbf{C}_b^n(\omega_{nb}^b \times), \quad \omega_{nb}^b = \omega_{ib}^b - \mathbf{b}_g - \mathbf{C}_n^b \omega_{ie}^n \quad (5)$$

$$\mathbf{C}_b^n(\mathbf{f}^b - \mathbf{b}_a) + \mathbf{g}^n = \mathbf{0} \quad (6)$$

$$\dot{\mathbf{b}}_g = \mathbf{0}, \quad \dot{\mathbf{b}}_a = \mathbf{0} \quad (7)$$

where the initial state includes the initial attitude matrix $\mathbf{C}_n^b(0)$, the gyroscope drift $\mathbf{b}_g$ and the accelerometer bias $\mathbf{b}_a$ Unlike a linear system, whose observability is irrelevant to the system input, observability of a nonlinear system highly depends on the system input [11, 13]. In this case, system input refers to the body angular rate $\omega_{ib}^b$ and the specific force $\mathbf{f}^b$. Equivalently, the SAOP investigates the effect of the known body angular rate and specific force on state observability. The required body angular rate and specific force are fulfilled by attitude motion.

## III. GLOBAL PERSPECTIVE OF OBSERVABILITY

This section shows how to attack the SAOP from a global perspective by decoupling and solving the nonlinear (5)–(7). If not explicitly stated, the strapdown INS is not located at the Earth's poles, i.e., $L \neq \pm \pi/2$.

First, we present several lemmas, which are used later.

LEMMA 1 [37, 38] *For any two linearly independent vectors, if their coordinates in two arbitrary frames are given, then the attitude matrix between the two frames can be determined.*

LEMMA 2 *Given $m$ known points $\mathbf{a}_k$, $k = 1, 2, \ldots, m$, in three-dimensional space satisfying $|\mathbf{a}_k - \mathbf{x}| = r$, where*

---

[3]If not explicitly stated, quantities in this paper are time dependent. The dependence on $t$ is omitted for clearer presentation.

**x** is an unknown point, r is a positive scalar, and $|\cdot|$ is the norm operator. If points $\mathbf{a}_k$ do not lie in any common plane, **x** has a unique solution. See Appendix Section A for the proof.

LEMMA 3  *Let $\mathbf{a}(t)$ and $\mathbf{b}(t)$ be known three-dimensional vectors on some time interval that satisfy $\mathbf{a}(t) \times \mathbf{m} = \mathbf{b}(t)$, where $\mathbf{m}$ is an unknown constant vector. If $\mathbf{a}(t)$ has nonconstant directions, then $\mathbf{m}$ can be uniquely solved. See Appendix Section B for the proof.*

LEMMA 4  *Let $\mathbf{a}$ and $\mathbf{b}$ be known three-dimensional vectors satisfying $\mathbf{a} \times \mathbf{m} = \mathbf{b}$ ($|\mathbf{a}| \neq 0$), where $\mathbf{m}$ is an unknown vector. If $|\mathbf{m}|$ is given, then $\mathbf{m}$ has solutions expressed as $\mathbf{m} = \pm \mathbf{a}\sqrt{|\mathbf{a}|^2|\mathbf{m}|^2 - |\mathbf{b}|^2}/|\mathbf{a}|^2 - \mathbf{a} \times \mathbf{b}/|\mathbf{a}|^2$. See Appendix Section C for the proof.*

### A. Static Alignment

For most applications, the strapdown INS has to align itself under a still condition, which is known as static alignment. This kind of alignment has been most frequently studied so far [1, 15, 16, 33].

THEOREM 1  *For static alignment, the SAOP is unobservable. The number of unobservable states is infinite.*

PROOF OF THEOREM 1  During the static period, the body angular rate $\boldsymbol{\omega}_{ib}^b$ and the specific force $\mathbf{f}^b$ are constants and $\boldsymbol{\omega}_{nb}^b = 0$. From (5), it gives

$$\boldsymbol{\omega}_{ib}^b - \mathbf{b}_g = \mathbf{C}_n^b \boldsymbol{\omega}_{ie}^n. \tag{8}$$

Attitude transformation does not change the magnitude of a vector, so we have

$$|\boldsymbol{\omega}_{ib}^b - \mathbf{b}_g| = \Omega \tag{9}$$

which means that the solution of $\mathbf{b}_g$ can be any point on the sphere surface with radius $\Omega$ that centers on $\boldsymbol{\omega}_{ib}^b$. Because $\boldsymbol{\omega}_{ib}^b$ is constant, (9) imposes a constraint on $\mathbf{b}_g$ that has three unknown components.

Similarly, taking norms on both sides of (6) indicates

$$|\mathbf{f}^b - \mathbf{b}_a| = g \tag{10}$$

where $g$ is the known magnitude of the local gravity. It shows that the solution of $\mathbf{b}_a$ can be any point of the sphere surface with radius $g$ that centers on $\mathbf{f}^b$. Equation (10) imposes a constraint on the three unknown components of $\mathbf{b}_a$.

The biases $\mathbf{b}_g$ and $\mathbf{b}_a$ are not independent. Equations (6) and (8) imply

$$(\boldsymbol{\omega}_{ib}^b - \mathbf{b}_g)^T (\mathbf{f}^b - \mathbf{b}_a) = -\boldsymbol{\omega}_{ie}^{nT} \mathbf{C}_b^n \mathbf{C}_n^b \mathbf{g}^n = g\Omega \sin L \tag{11}$$

which imposes one more constraint on $\mathbf{b}_g$ and $\mathbf{b}_a$. Using (9) and (10), we see that (11) says the angle formed by the two vectors $\boldsymbol{\omega}_{ib}^b - \mathbf{b}_g$ and $\mathbf{f}^b - \mathbf{b}_a$ is $\pi/2 - L$.

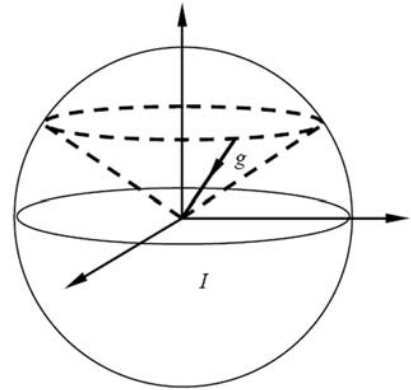

Fig. 1. Trajectory history of gravity vector in inertial frame depicts cone.

Designate an inertial frame $I$ to be the local-level frame at $t = 0$, i.e., $I = N(0)$. Decompose the body attitude matrix at the current time as

$$\mathbf{C}_n^b = \mathbf{C}_{b(0)}^{b(t)} \mathbf{C}_{n(0)}^{b(0)} \mathbf{C}_{n(t)}^{n(0)} = \mathbf{C}_{b(0)}^{b(t)} \mathbf{C}_n^b(0) \mathbf{C}_{n(t)}^{n(0)} \tag{12}$$

where $\mathbf{C}_{b(0)}^{b(t)}$ and $\mathbf{C}_{n(t)}^{n(0)}$ are attitude matrices as functions of $\boldsymbol{\omega}_{ib}^b - \mathbf{b}_g$ and $\boldsymbol{\omega}_{ie}^n$, respectively. They encode, respectively, the attitude changes of the body frame and the navigation frame from time 0 to $t$. Substituting into (6) gives

$$\mathbf{C}_n^b(0) \mathbf{C}_{n(t)}^{n(0)} \mathbf{g}^n = \mathbf{C}_{b(t)}^{b(0)} (\mathbf{b}_a - \mathbf{f}^b). \tag{13}$$

The quantity $\mathbf{C}_{n(t)}^{n(0)} \mathbf{g}^n$ is the gravity vector seen from the inertial frame $I$, and its trajectory history forms a cone at all locations but the two Earth poles where the cone degenerates to a line, as shown in Fig. 1. So there always exist two time instants that $\mathbf{C}_{n(t)}^{n(0)} \mathbf{g}^n$ have linearly independent directions. Using Lemma 1, the initial body attitude $\mathbf{C}_n^b(0)$ can be solved if only $\mathbf{b}_g$ and $\mathbf{b}_a$ are given.

Because $\mathbf{b}_g$ and $\mathbf{b}_a$ are not definite, $\mathbf{C}_n^b(0)$ is indeterminate and has a solution for each feasible pair of $\mathbf{b}_g$ and $\mathbf{b}_a$. Therefore, the system is unobservable, and the unobservable states are constrained by (9)–(11) and (13).

REMARK 1  Given $\mathbf{b}_g$ and $\mathbf{b}_a$, the initial body matrix $\mathbf{C}_n^b(0)$ can be analytically solved. Right multiplying $(\mathbf{C}_{n(t)}^{n(0)} \mathbf{g}^n)^T$ on both sides of and integrating over $[0, t_1]$, we have

$$\mathbf{C}_n^b(0) \int_0^{t_1} \mathbf{C}_{n(t)}^{n(0)} \mathbf{g}^n \mathbf{g}^{nT} \mathbf{C}_{n(0)}^{n(t)} dt$$
$$= \int_0^{t_1} \mathbf{C}_{b(t)}^{b(0)} (\mathbf{b}_a - \mathbf{f}^b) \mathbf{g}^{nT} \mathbf{C}_{n(0)}^{n(t)} dt \tag{14}$$

from which we solve

$$\mathbf{C}_n^b(0) = \int_0^{t_1} \mathbf{C}_{b(t)}^{b(0)} (\mathbf{b}_a - \mathbf{f}^b) \mathbf{g}^{nT} \mathbf{C}_{n(0)}^{n(t)} dt$$
$$\cdot \left[ \int_0^{t_1} \mathbf{C}_{n(t)}^{n(0)} \mathbf{g}^n \mathbf{g}^{nT} \mathbf{C}_{n(0)}^{n(t)} dt \right]^{-1}. \tag{15}$$

REMARK 2  Theorem 1 shows that we cannot accomplish the state estimation during static alignment. If a full-state estimator, e.g., the Kalman filter, is designed to do the static alignment, the estimator is supposed to converge to one of the unobservable states depending on the estimator settings, e.g., the selection of initial value (see Static Alignment in Section IV).

In the linearization-based observability analysis, the state space was usually divided into the observable subspace and the unobservable subspace to achieve better accuracy of estimation [15, 16]. This was done by setting all but one unobservable state in the observable combination to zeros. In the nonlinear context given here, it is equivalent to imposing constraints in addition to (9)–(11) and (13). For a navigational strapdown INS, we may as well perform the static alignment by simply assuming zero inertial sensor biases. This is common sense for practitioners (e.g., see [39]). The standard deviations of the neglected sensor biases impose a limit on the accuracy to which the remaining states may be determined.

Theorem 1 can be generalized to portray the problem of multiposition alignment.

THEOREM 2  *Consider a strapdown INS over $[0,t_1]$. If it is static on several disconnected subintervals such that either $\omega_{ib}^b$ or $\mathbf{f}^b$ on these static segments does not lie in any common plane, then the system is observable.*

PROOF OF THEOREM 2  The determination of $\mathbf{b}_g$ and $\mathbf{b}_a$ can be achieved by using Lemma 2. Let us first consider $\omega_{ib}^b$. As for (9), we know from Lemma 2 that $\mathbf{b}_g$ has a unique solution if $\omega_{ib}^b$ on these static segments does not lie in any common plane. Consequently, the known $\omega_{ib}^b - \mathbf{b}_g$ does not lie in one plane as well. So $\mathbf{b}_a$ is uniquely determined, noting (11) can be rewritten as $(\omega_{ib}^b - \mathbf{b}_g)^T \mathbf{b}_a = (\omega_{ib}^b - \mathbf{b}_g)^T \mathbf{f}^b - g\Omega \sin L$. The same story goes if we start the discussion from $\mathbf{f}^b$. Once the correct $\mathbf{b}_g$ and $\mathbf{b}_a$ are found, the initial body matrix $\mathbf{C}_n^b(0)$ is analytically solved.

REMARK 3  For the multiposition alignment in [7], it was claimed (Theorem 2 therein) that two still positions with different heading angles result in an observable system. We can readily show using Theorem 2 and Lemma 2 that the constraints (9)–(11) introduced by such two still positions are not enough to guarantee unique solutions of the sensor biases. In fact, Lemma 2 says that at least four still positions are required. As shown later, the rotating motion between still positions matters for observability. As a counterexample, consider a rotating motion that never stops or stops at an end position with the same heading as the start position. Theorem 2 in [7] says nothing about this, although the defect was partly remedied by a later work [24] using Lyapunov transformation and by our short note [25].

Theorem 2 employs only the information during the disconnected stays and thus requests tight conditions for an observable system. It is attitude motion that brings the strapdown INS from one static position to another. We next investigate attitude motion's contribution to observability.

B. Tumbling Alignment

The proper way to obtain accurate alignment is to improve the observability by, e.g., maneuvering [7–10, 14, 18, 24, 40, 41]. Observability has a tight connection with the input for a nonlinear system [11, 13]. As far as the SAOP is concerned, the observability is affected by the input, i.e., the body angular rate and specific force. Constant-speed rotation is considered first and then extended to varying-speed rotation.

THEOREM 3  *Consider the system rotating with the non-zero constant $\omega_{nb}^b$ ($\omega_{nb}^b \neq 0$ and $\dot{\omega}_{nb}^b = 0$) over $[0,t_1]$. Suppose $\dot{\omega}_{ib}^b$ or $\dot{\mathbf{f}}^b$ has a nonconstant direction:*

1) *If $\omega_{nb}^b$ is not perpendicular to both $\mathbf{g}^b$ and $\omega_{ie}^b$, the system is unobservable with two indistinguishable states.*

2) *If $\omega_{nb}^b$ is perpendicular to $\mathbf{g}^b$ but not to $\omega_{ie}^b$, the system is unobservable with two indistinguishable states. In addition, $\mathbf{b}_a$ is observable.*

3) *If $\omega_{nb}^b$ is perpendicular to $\omega_{ie}^b$ but not to $\mathbf{g}^b$, the system is unobservable with two indistinguishable states. In addition, $\mathbf{b}_g$ is observable.*

4) *If $\omega_{nb}^b$ is perpendicular to both $\mathbf{g}^b$ and $\omega_{ie}^b$, the system is observable.*

PROOF OF THEOREM 3  From (5), we have

$$\omega_{nb}^b = \omega_{ib}^b - b_g - \mathbf{C}_n^b \omega_{ie}^n. \quad (16)$$

Because $\omega_{nb}^b$ is constant, taking the time derivative on both sides and substituting (5) and (7) gives

$$\mathbf{0} = \dot{\omega}_{ib}^b + (\omega_{nb}^b \times) \mathbf{C}_n^b \omega_{ie}^n. \quad (17)$$

With (16), the preceding equation is rewritten as

$$\mathbf{0} = \dot{\omega}_{ib}^b + \omega_{nb}^b \times (\omega_{ib}^b - b_g). \quad (18)$$

Taking the time derivative again on both sides, we obtain

$$\dot{\omega}_{ib}^b \times \omega_{nb}^b = \ddot{\omega}_{ib}^b. \quad (19)$$

In addition, the time derivative of (6) yields

$$\mathbf{C}_b^n (\omega_{nb}^b \times (f^b - b_a) + \dot{\mathbf{f}}^b) = \mathbf{0} \quad (20)$$

or, equivalently,

$$\omega_{nb}^b \times (f^b - b_a) + \dot{\mathbf{f}}^b = \mathbf{0}. \quad (21)$$

Taking the time derivative again, it gives

$$\dot{\mathbf{f}}^b \times \omega_{nb}^b = \ddot{\mathbf{f}}^b. \quad (22)$$

Because $\dot{\boldsymbol{\omega}}_{ib}^b$ or $\dot{\mathbf{f}}^b$ has a nonconstant direction as assumed, Lemma 3 tells us that $\boldsymbol{\omega}_{nb}^b$ can be uniquely determined from (19) or (22). Now $\boldsymbol{\omega}_{nb}^b$ can be used as a known quantity.

Rewrite (21) as

$$\boldsymbol{\omega}_{nb}^b \times (\mathbf{f}^b - \mathbf{b}_a) = -\dot{\mathbf{f}}^b. \quad (23)$$

According to (10), $|\mathbf{f}^b - \mathbf{b}_a| = g$. Using Lemma 4, the solution of $\mathbf{b}_a$ is

$$\mathbf{b}_{a+,-} = \mathbf{f}^b \pm \frac{\boldsymbol{\omega}_{nb}^b \sqrt{g^2|\boldsymbol{\omega}_{nb}^b|^2 - |\dot{\mathbf{f}}^b|^2}}{|\boldsymbol{\omega}_{nb}^b|^2} - \frac{\boldsymbol{\omega}_{nb}^b \times \dot{\mathbf{f}}^b}{|\boldsymbol{\omega}_{nb}^b|^2}. \quad (24)$$

Substituting (23) and using (6), it is reduced to

$$\mathbf{b}_{a+,-} = \mathbf{f}^b \pm \frac{\boldsymbol{\omega}_{nb}^b |\boldsymbol{\omega}_{nb}^b \cdot \mathbf{g}^b|}{|\boldsymbol{\omega}_{nb}^b|^2} - \frac{\boldsymbol{\omega}_{nb}^b \times \dot{\mathbf{f}}^b}{|\boldsymbol{\omega}_{nb}^b|^2}$$

$$= \mathbf{b}_a + \frac{\boldsymbol{\omega}_{nb}^b (\pm|\boldsymbol{\omega}_{nb}^b \cdot \mathbf{g}^b| - \boldsymbol{\omega}_{nb}^b \cdot \mathbf{g}^b)}{|\boldsymbol{\omega}_{nb}^b|^2} \quad (25)$$

where $\mathbf{g}^b = \mathbf{C}_n^b \mathbf{g}^n = \mathbf{b}_a - \mathbf{f}^b$ according to (6) and "·" denotes the dot product of vectors. When $\boldsymbol{\omega}_{nb}^b$ is perpendicular to $\mathbf{g}^b$, $\mathbf{b}_{a+,-} = \mathbf{b}_a$ and we have the correct solution; otherwise, we get two distinctive solutions, one of which is correct.

Rewrite (18) as

$$\boldsymbol{\omega}_{nb}^b \times (\boldsymbol{\omega}_{ib}^b - \mathbf{b}_g - \boldsymbol{\omega}_{nb}^b) = -\dot{\boldsymbol{\omega}}_{ib}^b. \quad (26)$$

According to (16), $|\boldsymbol{\omega}_{ib}^b - \mathbf{b}_g - \boldsymbol{\omega}_{nb}^b| = \Omega$. With the help of Lemma 4, the solution of $\mathbf{b}_g$ is given by

$$\mathbf{b}_{g+,-} = \boldsymbol{\omega}_{ib}^b - \boldsymbol{\omega}_{nb}^b \pm \frac{\boldsymbol{\omega}_{nb}^b \sqrt{\Omega^2|\boldsymbol{\omega}_{nb}^b|^2 - |\dot{\boldsymbol{\omega}}_{ib}^b|^2}}{|\boldsymbol{\omega}_{nb}^b|^2}$$

$$- \frac{\boldsymbol{\omega}_{nb}^b \times \dot{\boldsymbol{\omega}}_{ib}^b}{|\boldsymbol{\omega}_{nb}^b|^2}. \quad (27)$$

Substituting (26) and using (16), it yields

$$\mathbf{b}_{g+,-} = \boldsymbol{\omega}_{ib}^b - \boldsymbol{\omega}_{nb}^b \pm \frac{\boldsymbol{\omega}_{nb}^b |\boldsymbol{\omega}_{nb}^b \cdot \boldsymbol{\omega}_{ie}^b|}{|\boldsymbol{\omega}_{nb}^b|^2} - \frac{\boldsymbol{\omega}_{nb}^b \times \dot{\boldsymbol{\omega}}_{ib}^b}{|\boldsymbol{\omega}_{nb}^b|^2}$$

$$= \mathbf{b}_g + \frac{\boldsymbol{\omega}_{nb}^b(\pm|\boldsymbol{\omega}_{nb}^b \cdot \boldsymbol{\omega}_{ie}^b| + \boldsymbol{\omega}_{nb}^b \cdot \boldsymbol{\omega}_{ie}^b)}{|\boldsymbol{\omega}_{nb}^b|^2} \quad (28)$$

where $\boldsymbol{\omega}_{ie}^b = \mathbf{C}_n^b \boldsymbol{\omega}_{ie}^n = \boldsymbol{\omega}_{ib}^b - \mathbf{b}_g - \boldsymbol{\omega}_{nb}^b$, according to (16). When $\boldsymbol{\omega}_{nb}^b$ is perpendicular to $\boldsymbol{\omega}_{ie}^b$, $\mathbf{b}_{g+,-} = \mathbf{b}_g$ and it gives the correct solution; otherwise, we get two distinctive solutions, one of which is correct.

For each feasible $(\mathbf{b}_a, \mathbf{b}_g)$ pair, the corresponding solution of the initial attitude matrix $\mathbf{C}_n^b(0)$ is given by (15). Therefore, the observability result depends on the following:

1) $\boldsymbol{\omega}_{nb}^b$ is not perpendicular to both $\mathbf{g}^b$ and $\boldsymbol{\omega}_{ie}^b$. Both $\mathbf{b}_a$ and $\mathbf{b}_g$ have two solutions. We have four possible $(\mathbf{b}_a, \mathbf{b}_g)$ pairs, among which only two are valid (see Appendix Section D for explanations), so the system is unobservable with two indistinguishable states. Specifically, if $(\boldsymbol{\omega}_{nb}^b \cdot \boldsymbol{\omega}_{ie}^b) \cdot (\boldsymbol{\omega}_{nb}^b \cdot \mathbf{g}^b) < 0$, the feasible pairs should be $(\mathbf{b}_{a+}, \mathbf{b}_{g+})$ and $(\mathbf{b}_{a-}, \mathbf{b}_{g-})$; otherwise, the feasible pairs are $(\mathbf{b}_{a+}, \mathbf{b}_{g-})$ and $(\mathbf{b}_{a-}, \mathbf{b}_{g+})$.

2) $\boldsymbol{\omega}_{nb}^b$ is perpendicular to $\mathbf{g}^b$ but not to $\boldsymbol{\omega}_{ie}^b$. $\mathbf{b}_a$ has one solution, but $\mathbf{b}_g$ has two solutions. There are two feasible $(\mathbf{b}_a, \mathbf{b}_g)$ pairs, so the system is unobservable with two indistinguishable states.

3) $\boldsymbol{\omega}_{nb}^b$ is perpendicular to $\boldsymbol{\omega}_{ie}^b$ but not to $\mathbf{g}^b$. $\mathbf{b}_a$ has two solutions and $\mathbf{b}_g$ has one solution. We have two feasible $(\mathbf{b}_a, \mathbf{b}_g)$ pairs, and the system is unobservable with two indistinguishable states.

4) $\boldsymbol{\omega}_{nb}^b$ is perpendicular to both $\mathbf{g}^b$ and $\boldsymbol{\omega}_{ie}^b$. Both $\mathbf{b}_a$ and $\mathbf{b}_g$ have one solution, so the system is observable.

REMARK 4  The precondition "$\dot{\boldsymbol{\omega}}_{ib}^b$ or $\dot{\mathbf{f}}^b$ has a nonconstant direction" is not easy to check because of derivative involvement. Equation (26) shows that $\dot{\boldsymbol{\omega}}_{ib}^b$ is perpendicular to $\boldsymbol{\omega}_{nb}^b$. With (26) and (16), we have

$$|\dot{\boldsymbol{\omega}}_{ib}^b|^2 = |\boldsymbol{\omega}_{nb}^b \times \boldsymbol{\omega}_{ie}^b|^2. \quad (29)$$

Using (19), we obtain

$$\dot{\boldsymbol{\omega}}_{ib}^b \times \ddot{\boldsymbol{\omega}}_{ib}^b = \dot{\boldsymbol{\omega}}_{ib}^b \times (\dot{\boldsymbol{\omega}}_{ib}^b \times \boldsymbol{\omega}_{nb}^b) = -|\dot{\boldsymbol{\omega}}_{ib}^b|^2 \boldsymbol{\omega}_{nb}^b$$

$$= -|\boldsymbol{\omega}_{nb}^b \times \boldsymbol{\omega}_{ie}^b|^2 \boldsymbol{\omega}_{nb}^b. \quad (30)$$

That "$\dot{\boldsymbol{\omega}}_{ib}^b$ has a nonconstant direction" ($\dot{\boldsymbol{\omega}}_{ib}^b \times \ddot{\boldsymbol{\omega}}_{ib}^b \neq 0$) is identical to "$\boldsymbol{\omega}_{nb}^b$ is unparallel to $\boldsymbol{\omega}_{ie}^b$." Similarly, (23) shows that $\dot{\mathbf{f}}^b$ is perpendicular to $\boldsymbol{\omega}_{nb}^b$. With (23) and (6), it yields

$$|\dot{\mathbf{f}}^b|^2 = |\boldsymbol{\omega}_{nb}^b \times \mathbf{g}^b|^2. \quad (31)$$

With (22),

$$\dot{\mathbf{f}}^b \times \ddot{\mathbf{f}}^b = \dot{\mathbf{f}}^b \times (\dot{\mathbf{f}}^b \times \boldsymbol{\omega}_{nb}^b) = -|\dot{\mathbf{f}}^b|^2 \boldsymbol{\omega}_{nb}^b = -|\boldsymbol{\omega}_{nb}^b \times \mathbf{g}^b|^2 \boldsymbol{\omega}_{nb}^b \quad (32)$$

It shows that "$\dot{\mathbf{f}}^b$ has a nonconstant direction" ($\dot{\mathbf{f}}^b \times \ddot{\mathbf{f}}^b \neq 0$) is identical to "$\boldsymbol{\omega}_{nb}^b$ is unparallel to $\mathbf{g}^b$."

Theorem 3 makes moderate assumptions and has a wide scope of applicability. Consider the case in which $\dot{\boldsymbol{\omega}}_{ib}^b$ and $\dot{\mathbf{f}}^b$ have constant directions in the entire interval. In such a case, (30) and (32) show that the three vectors $\boldsymbol{\omega}_{nb}^b$, $\boldsymbol{\omega}_{ie}^b$, and $\mathbf{g}^b$ must be in the same direction. It refers to a strapdown INS system at $L = \pm\pi/2$ rotating with respect to the Earth along the Earth's axis. This case is extremely rare.

REMARK 5  The indistinguishable states $\mathbf{b}_g$ and $\mathbf{b}_a$ are separated from each other in the direction of $\boldsymbol{\omega}_{nb}^b$. As shown by (25) and (28), $\mathbf{b}_{a+} - \mathbf{b}_{a-} = 2\boldsymbol{\omega}_{nb}^b|\boldsymbol{\omega}_{nb}^b \cdot \mathbf{g}^b|/|\boldsymbol{\omega}_{nb}^b|^2$ and $\mathbf{b}_{g+} - \mathbf{b}_{g-} = 2\boldsymbol{\omega}_{nb}^b|\boldsymbol{\omega}_{nb}^b \cdot \boldsymbol{\omega}_{ie}^b|/|\boldsymbol{\omega}_{nb}^b|^2$, so we have $|\mathbf{b}_{a+} - \mathbf{b}_{a-}| = 2g|\cos(\widehat{\boldsymbol{\omega}_{nb}^b, \mathbf{g}^b})|$ and $|\mathbf{b}_{g+} - \mathbf{b}_{g-}| = 2\Omega|\cos(\widehat{\boldsymbol{\omega}_{nb}^b, \boldsymbol{\omega}_{ie}^b})|$, where $\cos(\widehat{\cdot,\cdot})$ denotes the cosine

of the angle formed by two vectors. The distances between the indistinguishable states of $\mathbf{b}_g$ and $\mathbf{b}_a$ depend on the relation of $\boldsymbol{\omega}_{nb}^b$ with respect to $\mathbf{g}^b$ and $\boldsymbol{\omega}_{ie}^b$, respectively.

REMARK 6  $\mathbf{g}^b$ is in the vertical direction, and $\boldsymbol{\omega}_{ie}^b$ is parallel to the Earth axis. The following are natural corollaries of Theorem 3: 1) If $\boldsymbol{\omega}_{nb}^b$ points in the local vertical direction, the system has two indistinguishable states. In addition, $\mathbf{b}_g$ is observable when the system is located at the equator ($L = 0$). 2) If $\boldsymbol{\omega}_{nb}^b$ points in the north-south direction, the system has two indistinguishable states. In addition, $\mathbf{b}_a$ is observable. 3) If $\boldsymbol{\omega}_{nb}^b$ points in the east-west direction, the system is observable.

The following theorem shows that it is not the static positions but the rotating motion between that matters for observability. In other words, static segments theoretically contribute nothing to observability improvement.

THEOREM 4  *Consider the system over $[0, t_1]$. It rotates with the non-zero constant $\boldsymbol{\omega}_{nb}^b$ over the subinterval $[t_2, t_3] \in [0, t_1]$ and stays static for the other periods. If $\dot{\boldsymbol{\omega}}_{ib}^b$ or $\dot{\mathbf{f}}^b$ has a nonconstant direction over $[t_2, t_3]$, then the claims are the same as in Theorem 3.*

PROOF OF THEOREM 4  Because Theorem 3 is directly applicable to the subinterval $[t_2, t_3]$, the claim will be proved if we can show that $\mathbf{b}_{g+,-}$ and $\mathbf{b}_{a+,-}$ satisfy (9)–(11) during the static periods. Without loss of generality, we assume $\mathbf{b}_{g+} = \mathbf{b}_g$ ($\boldsymbol{\omega}_{nb}^b \cdot \boldsymbol{\omega}_{ie}^b < 0$) and $\mathbf{b}_{a+} = \mathbf{b}_a$ ($\boldsymbol{\omega}_{nb}^b \cdot \mathbf{g}^b > 0$). The correct biases $\mathbf{b}_{g+}$ and $\mathbf{b}_{a+}$ naturally satisfy (9)–(11). Then we have

$$|\boldsymbol{\omega}_{ib}^b(t_s) - \mathbf{b}_{g-}|^2 = |\boldsymbol{\omega}_{ib}^b(t_s) - \mathbf{b}_{g+} + \mathbf{b}_{g+} - \mathbf{b}_{g-}|^2$$
$$= \Omega^2 + 2(\boldsymbol{\omega}_{ib}^b(t_s) - \mathbf{b}_{g+})^T(\mathbf{b}_{g+} - \mathbf{b}_{g-})$$
$$+ |\mathbf{b}_{g+} - \mathbf{b}_{g-}|^2 \quad (33)$$

where $\boldsymbol{\omega}_{ib}^b(t_s)$ denotes the measured body angular rate when the system is in static condition at $t_s$. From (8), $\boldsymbol{\omega}_{ib}^b(t_s) - \mathbf{b}_{g+} = \boldsymbol{\omega}_{ie}^b(t_s)$. Substituting into (33), we get from Remark 5

$$|\boldsymbol{\omega}_{ib}^b(t_s) - \mathbf{b}_{g-}|^2$$
$$= \Omega^2 + 4\frac{(\boldsymbol{\omega}_{nb}^b \cdot \boldsymbol{\omega}_{ie}^b(t_s))|\boldsymbol{\omega}_{nb}^b \cdot \boldsymbol{\omega}_{ie}^b|}{|\boldsymbol{\omega}_{nb}^b|^2} + 4\Omega^2\cos^2(\widehat{\boldsymbol{\omega}_{nb}^b, \boldsymbol{\omega}_{ie}^b})$$
$$= \Omega^2 + 4\frac{(\boldsymbol{\omega}_{nb}^b \cdot \boldsymbol{\omega}_{ie}^b)|\boldsymbol{\omega}_{nb}^b \cdot \boldsymbol{\omega}_{ie}^b|}{|\boldsymbol{\omega}_{nb}^b|^2} + 4\Omega^2\cos^2(\widehat{\boldsymbol{\omega}_{nb}^b, \boldsymbol{\omega}_{ie}^b})$$
$$\stackrel{\boldsymbol{\omega}_{nb}^b \cdot \boldsymbol{\omega}_{ie}^b < 0}{=} \Omega^2 - 4\frac{|\boldsymbol{\omega}_{nb}^b \cdot \boldsymbol{\omega}_{ie}^b|^2}{|\boldsymbol{\omega}_{nb}^b|^2} + 4\Omega^2\cos^2(\widehat{\boldsymbol{\omega}_{nb}^b, \boldsymbol{\omega}_{ie}^b})$$
$$= \Omega^2 \quad (34)$$

where the second equality is valid because the physical vector $\boldsymbol{\omega}_{nb}$ is unchanged with respect to the Earth, i.e.,

$$\widehat{\boldsymbol{\omega}_{nb}^b(t), \boldsymbol{\omega}_{ie}^b(t_s)} = \widehat{\boldsymbol{\omega}_{nb}^b(t_2), \boldsymbol{\omega}_{ie}^b(t_2)} = \widehat{\boldsymbol{\omega}_{nb}^b(t), \boldsymbol{\omega}_{ie}^b(t)}$$
$$\text{for} \quad t_s < t_2$$
$$\widehat{\boldsymbol{\omega}_{nb}^b(t), \boldsymbol{\omega}_{ie}^b(t_s)} = \widehat{\boldsymbol{\omega}_{nb}^b(t_3), \boldsymbol{\omega}_{ie}^b(t_3)} = \widehat{\boldsymbol{\omega}_{nb}^b(t), \boldsymbol{\omega}_{ie}^b(t)} \quad (35)$$
$$\text{for} \quad t_s > t_3.$$

Similarly, we have

$$|\mathbf{f}^b(t_s) - \mathbf{b}_{a-}|^2 = |\mathbf{f}^b(t_s) - \mathbf{b}_{a+} + \mathbf{b}_{a+} - \mathbf{b}_{a-}|^2$$
$$= g^2 + 2(\mathbf{f}^b(t_s) - \mathbf{b}_{a+})^T(\mathbf{b}_{a+} - \mathbf{b}_{a-})$$
$$+ |\mathbf{b}_{a+} - \mathbf{b}_{a-}|^2 \quad (36)$$

where $\mathbf{f}^b(t_s)$ is the measured specific force when the system is static. From (6), $\mathbf{f}^b(t_s) - \mathbf{b}_{a+} = -\mathbf{g}^b(t_s)$ Substituting into (36), we obtain from Remark 5

$$|\mathbf{f}^b(t_s) - \mathbf{b}_{a-}|^2$$
$$= g^2 - 4\frac{(\boldsymbol{\omega}_{nb}^b \cdot \mathbf{g}^b(t_s))|\boldsymbol{\omega}_{nb}^b \cdot \mathbf{g}^b|}{|\boldsymbol{\omega}_{nb}^b|^2} + 4g^2\cos^2(\widehat{\boldsymbol{\omega}_{nb}^b, \mathbf{g}^b})$$
$$= g^2 - 4\frac{(\boldsymbol{\omega}_{nb}^b \cdot \mathbf{g}^b)|\boldsymbol{\omega}_{nb}^b \cdot \mathbf{g}^b|}{|\boldsymbol{\omega}_{nb}^b|^2} + 4g^2\cos^2(\widehat{\boldsymbol{\omega}_{nb}^b, \mathbf{g}^b})$$
$$\stackrel{\boldsymbol{\omega}_{nb}^b \cdot \mathbf{g}^b > 0}{=} g^2 - 4\frac{|\boldsymbol{\omega}_{nb}^b \cdot \mathbf{g}^b|^2}{|\boldsymbol{\omega}_{nb}^b|^2} + 4g^2\cos^2(\widehat{\boldsymbol{\omega}_{nb}^b, \mathbf{g}^b})$$
$$= g^2. \quad (37)$$

As far as (11) is concerned, it yields

$$(\boldsymbol{\omega}_{ib}^b(t_s) - \mathbf{b}_{g-})^T(\mathbf{f}^b(t_s) - \mathbf{b}_{a-})$$
$$= (\boldsymbol{\omega}_{ib}^b(t_s) - \mathbf{b}_{g+} + \mathbf{b}_{g+} - \mathbf{b}_{g-})^T$$
$$\cdot (\mathbf{f}^b(t_s) - \mathbf{b}_{a+} + \mathbf{b}_{a+} - \mathbf{b}_{a-})$$
$$= g\Omega\sin L + \boldsymbol{\omega}_{ie}^{bT}(t_s)(\mathbf{b}_{a+} - \mathbf{b}_{a-})$$
$$- (\mathbf{b}_{g+} - \mathbf{b}_{g-})^T\mathbf{g}^b(t_s) + (\mathbf{b}_{g+} - \mathbf{b}_{g-})^T(\mathbf{b}_{a+} - \mathbf{b}_{a-})$$
$$= g\Omega\sin L + \frac{2(\boldsymbol{\omega}_{nb}^b \cdot \boldsymbol{\omega}_{ie}^b(t_s))|\boldsymbol{\omega}_{nb}^b \cdot \mathbf{g}^b| - 2(\boldsymbol{\omega}_{nb}^b \cdot \mathbf{g}^b(t_s))}{|\boldsymbol{\omega}_{nb}^b \cdot \boldsymbol{\omega}_{ie}^b| + 4|\boldsymbol{\omega}_{nb}^b \cdot \mathbf{g}^b||\boldsymbol{\omega}_{nb}^b \cdot \boldsymbol{\omega}_{ie}^b|}$$
$$\phantom{= g\Omega\sin L + }\overline{\phantom{|\boldsymbol{\omega}_{nb}^b|^2}}$$
$$\stackrel{\boldsymbol{\omega}_{nb}^b \cdot \boldsymbol{\omega}_{ie}^b < 0, \boldsymbol{\omega}_{nb}^b \cdot \mathbf{g}^b > 0}{=} g\Omega\sin L. \quad (38)$$

REMARK 7  For multiple static segments and rotating segments interlaced, the proof of Theorem 4 indicates that static segments contribute nothing to observability enhancement because their immediate neighboring rotating segments impose tighter constraints on the state. The previous results in [7], [24], and [25] contradict Theorem 4 and Remark 6 (see Table I for a simple comparison). For example, it was claimed there that if the strapdown INS is rotated around the vertical direction for some time, the alignment becomes observable and consequently globally observable (by definition). As discussed in Remark 3 and confirmed in the next section, the previous claims in [7], [24], and [25] are theoretically incorrect (the estimator may converge to a wrong solution for some initial value). The inconsistency of the linearization-based observability result occurs because of the weak points outlined in the Introduction.

TABLE I
Simple Comparison with Linearization-Based Observability Result

| Rotation Axis | Global Observability | Linearization-Based Observability |
|---|---|---|
| Vertical | Unobservable | Observable |
| North-South | Unobservable | Observable |
| East-West | Observable | Unobservable |

Theorem 3 is constructive in that its proof not only tells us whether the system is observable or not under the assumptions but also gives us the explicit form of analytical solutions to the observable or unobservable states. Using the constructive proofs, we can design an ideal observer to estimate the states. Here, the term "ideal" is used because it requires the exact first and second derivatives of gyroscope and accelerometer outputs.

Consider that the strapdown INS rotates at the non-zero constant $\boldsymbol{\omega}_{nb}^b$ over $[t_2, t_3]$. Rewrite (19) and (22) in a compact form as

$$\begin{bmatrix} \dot{\boldsymbol{\omega}}_{ib}^b \times \\ \dot{\mathbf{f}}^b \times \end{bmatrix} \boldsymbol{\omega}_{nb}^b = \begin{bmatrix} \ddot{\boldsymbol{\omega}}_{ib}^b \\ \ddot{\mathbf{f}}^b \end{bmatrix}, \qquad t \in [t_2, t_3]. \quad (39)$$

Left multiplying $[\dot{\boldsymbol{\omega}}_{ib}^b \times \quad \dot{\mathbf{f}}^b \times]$ on both sides and integrating over $[t_2, t_3]$ give

$$\int_{t_2}^{t_3} ((\dot{\boldsymbol{\omega}}_{ib}^b \times)^2 + (\dot{\mathbf{f}}^b \times)^2) dt \cdot \boldsymbol{\omega}_{nb}^b$$
$$= \int_{t_2}^{t_3} [\dot{\boldsymbol{\omega}}_{ib}^b \times \ddot{\boldsymbol{\omega}}_{ib}^b + \dot{\mathbf{f}}^b \times \ddot{\mathbf{f}}^b] dt \quad (40)$$

from which we can solve

$$\boldsymbol{\omega}_{nb}^b = \left[ \int_{t_2}^{t_3} ((\dot{\boldsymbol{\omega}}_{ib}^b \times)^2 + (\dot{\mathbf{f}}^b \times)^2) dt \right]^{-1}$$
$$\cdot \int_{t_2}^{t_3} [\dot{\boldsymbol{\omega}}_{ib}^b \times \ddot{\boldsymbol{\omega}}_{ib}^b + \dot{\mathbf{f}}^b \times \ddot{\mathbf{f}}^b] dt. \quad (41)$$

According to Lemma 3, the matrix inverse exists if $\dot{\boldsymbol{\omega}}_{ib}^b$ or $\dot{\mathbf{f}}^b$ has a nonconstant direction over $[t_2, t_3]$.

The ideal observer for non-zero constant rotation (IO-NCR) are summarized here for clarity:

1) To determine $\boldsymbol{\omega}_{nb}^b$ by (41).
2) To obtain the possible solutions of gyroscope and accelerometer biases by (24) and (27).
3) To obtain the initial body matrix by (15) for each feasible $(\mathbf{b}_a, \mathbf{b}_g)$ pair.

The major assumption of Theorems 3 and 4 (fixed-direction, fix-magnitude rotation) can be largely relaxed to allow for fixed-direction, varying-magnitude rotation. This has the practical significance of eliminating the smooth requirement of tumbling tables.

THEOREM 5 *Consider the system rotating with the non-zero, fixed-direction, varying-magnitude $\boldsymbol{\omega}_{nb}^b$ ($\dot{\boldsymbol{\omega}}_{nb}^b \neq 0$ and $\boldsymbol{\omega}_{nb}^b \times \dot{\boldsymbol{\omega}}_{nb}^b = 0$) over $[0, t_1]$. If $\dot{\mathbf{f}}^b$ is the non-zero ($\dot{\mathbf{f}}^b \neq 0$), then the claims are the same as in Theorem 3.*

PROOF OF THEOREM 5  As in the proof of Theorem 3, we obtain, using the time derivative of (6),

$$\dot{\mathbf{f}}^b = (\mathbf{f}^b - \mathbf{b}_a) \times \boldsymbol{\omega}_{nb}^b. \quad (42)$$

Taking the time derivative of (42) gives

$$\ddot{\mathbf{f}}^b = (\mathbf{f}^b - \mathbf{b}_a) \times \dot{\boldsymbol{\omega}}_{nb}^b + \dot{\mathbf{f}}^b \times \boldsymbol{\omega}_{nb}^b. \quad (43)$$

Because $\boldsymbol{\omega}_{nb}^b$ is parallel to $\dot{\boldsymbol{\omega}}_{nb}^b$ as assumed, (42) and (43) indicate that $\boldsymbol{\omega}_{nb}^b$ is normal to both $\dot{\mathbf{f}}^b$ and $\ddot{\mathbf{f}}^b$. That is, $\boldsymbol{\omega}_{nb}^b$ can be expressed as

$$\boldsymbol{\omega}_{nb}^b = k(t) \ddot{\mathbf{f}}^b \times \dot{\mathbf{f}}^b \quad (44)$$

where $k(t)$ is a scalar time function. Using (42)–(44), we obtain

$$\ddot{\mathbf{f}}^b \times \dot{\mathbf{f}}^b = (\dot{\mathbf{f}}^b \times \boldsymbol{\omega}_{nb}^b) \times \dot{\mathbf{f}}^b = |\dot{\mathbf{f}}^b|^2 \boldsymbol{\omega}_{nb}^b = k(t) |\dot{\mathbf{f}}^b|^2 \ddot{\mathbf{f}}^b \times \dot{\mathbf{f}}^b \quad (45)$$

from which we solve $k(t) = 1/|\dot{\mathbf{f}}^b|^2$, because $|\dot{\mathbf{f}}^b| \neq 0$. So,

$$\boldsymbol{\omega}_{nb}^b = \ddot{\mathbf{f}}^b \times \dot{\mathbf{f}}^b / |\dot{\mathbf{f}}^b|^2 \quad (46)$$

is known. The remaining part of the proof is almost the same as in Theorem 3 and thus has been omitted.

THEOREM 6  *Consider the system over $[0, t_1]$. It rotates with the non-zero, fixed-direction, varying-magnitude $\boldsymbol{\omega}_{nb}^b$ ($\boldsymbol{\omega}_{nb}^b \neq 0$ and $\boldsymbol{\omega}_{nb}^b \times \dot{\boldsymbol{\omega}}_{nb}^b = 0$) over the subinterval $[t_2, t_3]$ and stays static for the other periods. If $\dot{\mathbf{f}}^b$ is the non-zero ($\dot{\mathbf{f}}^b \neq 0$) over $[t_2, t_3]$, then the claims are the same as in Theorem 3.*

PROOF OF THEOREM 6  See the proof of Theorem 4.

By assuming $\dot{\mathbf{f}}^b \neq 0$, Theorems 5 and 6 are inapplicable when $\boldsymbol{\omega}_{nb}^b$ is in the vertical direction (see (31) and Remark 5). Consider that the strapdown INS rotates with the fixed-direction, varying-magnitude $\boldsymbol{\omega}_{nb}^b$ over $[t_2, t_3]$. We can construct a second ideal observer by the proof of Theorem 5, the ideal observer for non-zero, fixed-direction, varying-magnitude rotation (IO-NFVR)

1) To determine $\boldsymbol{\omega}_{nb}^b$ by (46).
2) To obtain the possible solutions of gyroscope and accelerometer biases by (24) and (27).
3) To obtain the initial body matrix by (15) for each feasible $(\mathbf{b}_a, \mathbf{b}_g)$ pair.

The following theorems prove that an observable SAOP can always be attained by successive tumbling around two or more different directions.

THEOREM 7 *If there are no less than two subintervals over $[0,t_1]$ on which $\omega_{nb}^b$ not only is a non-zero constant but also is linearly independent, then the system is observable.*

PROOF OF THEOREM 7 For each subinterval, $\omega_{nb}^b$ is a non-zero constant, so we have from (18) and (21) that

$$\omega_{nb}^b \times \mathbf{b}_g = \dot{\omega}_{ib}^b + \omega_{nb}^b \times \omega_{ib}^b$$
$$\omega_{nb}^b \times \mathbf{b}_a = \dot{\mathbf{f}}^b + \omega_{nb}^b \times \mathbf{f}^b \quad (47)$$

in which $\omega_{nb}^b$ is a known quantity. Because $\omega_{nb}^b$ on the subintervals is linearly independent as assumed, $\mathbf{b}_g$ and $\mathbf{b}_a$ can be determined by (47). Consequently, $\mathbf{C}_n^b(0)$ is unique.

THEOREM 8 *If there are no less than two subintervals over $[0,t_1]$ on which $\omega_{nb}^b$ not only has a non-zero, fixed-direction, varying-magnitude but also is linearly independent, and additionally $\dot{\mathbf{f}}^b \neq 0$, then the system is observable.*

PROOF OF THEOREM 8 The proof is almost the same as that for Theorem 7, except the first equation in (47) is replaced with

$$\omega_{nb}^b \times \mathbf{b}_g = \dot{\omega}_{ib}^b + \omega_{nb}^b \times \omega_{ib}^b - \dot{\omega}_{nb}^b. \quad (48)$$

REMARK 8 Theorems 2, 7, and 8 give conditions to obtain an observable system. Theorem 2 seems to be of little use in that static segments contribute nothing to observability, but it helps us when it comes to numerical computation. To solve the unknown state, Theorem 2 needs only the raw sensor outputs, while Theorems 7 and 8 require their derivative information, which magnifies the noises in practice. The difference relates to the so-called observability degree in some literature [7, 24]. The observability degree lacks a solid analytical basis because of its origin from numerical computation, but we find in our later simulations that static segments do help suppress the numerical errors.

IV. SIMULATION STUDY

This section is devoted to numerical verification of analytical results, using extensive simulations with the two ideal observers (IO-NCR and IO-NFVR) and a practical approximate nonlinear observer (extended Kalman filter, or EKF). The strapdown INS (gyroscope drift of 0.01 deg/h, accelerometer bias of 50 $\mu$g, and output bandwidth of 100 Hz) is assumed to be located at $L = 28.2204$ deg and $h = 60$ m. For the apparent exhibition of observability changes, the sensor noises are not added. Euler angles from the reference frame to the body frame are defined as first around the y-axis (yaw, $\psi$), followed by the z-axis (pitch, $\theta$) and then by the x-axis (roll, $\varphi$). The indirect form of EKF is used, and the linear error dynamic equation for the system equations in (1)–(4) is readily available in [1–3]. EKF has 12 states: 3 for angle error, 3 for velocity error, 3 for gyroscope bias error, and 3 for accelerometer bias error. If not explicitly stated, the following EKF settings are used: The first 20 s are for coarse alignment, and random error of 1 deg ($1\sigma$) is intentionally added to the coarse alignment result. The initial sensor biases are zeros.

A. Static Alignment

The strapdown INS initial true attitude is set to $\varphi = 20$ deg, $\psi = 30$ deg, and $\theta = 10$ deg. The alignment lasts for 300 s. In all simulations, EKF converges to different estimates depending on the initial attitude angle, as predicted in Remark 2. Results for two runs of simulation are given for demonstration. Figs. 2 and 3 present the estimated attitude and sensor bias, respectively, for the initial angle $[19.896 \text{ deg} \quad 29.626 \text{ deg} \quad 9.620 \text{ deg}]^T$, and Figs. 4 and 5 are for the other initial angle $[20.124 \text{ deg} \quad 31.123 \text{ deg} \quad 10.474 \text{ deg}]^T$. The filter converges quickly in both runs, but the two estimates are quite different and there seems no connection between them. In Fig. 6, we plot $|\omega_{ib}^b - \mathbf{b}_g|$, $|\mathbf{f}^b - \mathbf{b}_a|$, and $(\omega_{ib}^b - \mathbf{b}_g)^T(\mathbf{f}^b - \mathbf{b}_a)$ using the estimated sensor biases in the two runs, along with their analytical values from (9)–(11). The constraints equations shown in (9)–(11) clearly dictate how the unobservable sensor biases behave in the estimation process. The same goes for the attitude estimate, with (13) as the constraint.

Because sensor biases are relatively small in magnitude compared to gyroscope and accelerometer outputs, (11) approximately leads to $(\omega_{ib}^b - \mathbf{b}_g)^T \mathbf{f}^b \approx (\mathbf{f}^b - \mathbf{b}_a)^T \omega_{ib}^b \approx g\Omega \sin L$. This means that $\mathbf{b}_g$ lies on a circle, the intersection of the sphere surface $|\mathbf{b}_g - \omega_{ib}^b|$ and a cone, with $-\mathbf{f}^b$ being its rotation axis and the half angle $\pi/2 - L$; $\mathbf{b}_a$ lies on a circle, the intersection of the sphere surface $|\mathbf{b}_a - \mathbf{f}^b|$ and a cone, with $-\omega_{ib}^b$ being its rotation axis and the half angle $\pi/2 - L$. The vertexes of the two cones coincide with the centers of the two spheres. The two intersecting circles are the unobservable spaces of sensor biases. Fig. 7 illustrates the unobservable spaces by plotting the bias estimates of 1000 Monte Carlo EKF runs as dots in space (the random error added to coarse alignment increases to 5 deg, or $1\sigma$, for better visual effect). In the left graph, the gyroscope bias estimates apparently form a segment of a circle on the sphere surface $|\mathbf{b}_g - \omega_{ib}^b|$; the accelerometer bias estimates in the right graph, lying on the sphere surface $|\mathbf{b}_a - \mathbf{f}^b|$, are too close to be distinguished from one another.

B. Tumbling Alignment by Single-Axis Rotation

The strapdown INS initial true attitude angles are set to zeros for clear demonstration. Rotations along the three body axes are inspected in turn. The body axes are now in the local north, up, and east directions. As for EKF, the random error added to coarse alignment deceases to 0.1 deg (1$\sigma$) to shorten the transient stage.

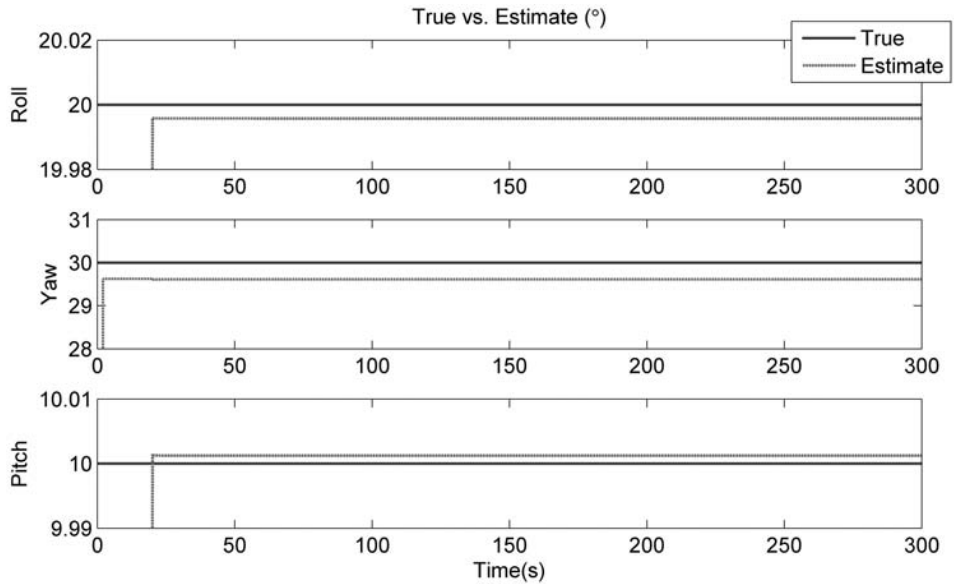

Fig. 2. Attitude estimates for initial angle [19.896 deg 29.626 deg 9.620 deg]$^T$.

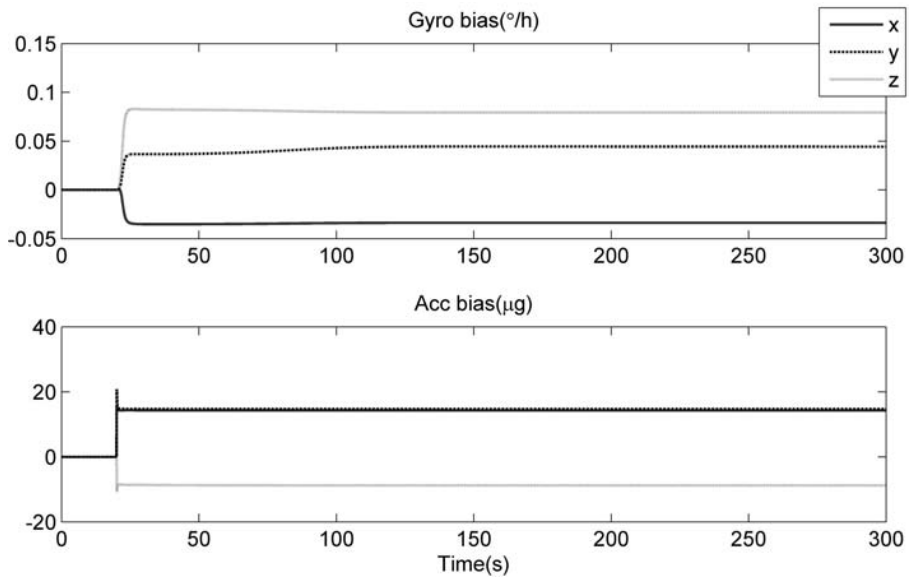

Fig. 3. Sensor bias estimates for initial angle [19.896 deg 29.626 deg 9.620 deg]$^T$.

*1) Up-Down Direction*: The strapdown INS is rotated along the up-down direction at 10 deg/s, i.e., $\omega_{nb}^b = [0, 10 \text{ deg/s}, 0]^T$. The alignment lasts up to 600 s. The rotation starts at 100 s and ends at 500 s. IO-NCR is applied to the rotating segment, and the bias estimates are given in Fig. 8 (zeros for 0–100 s and 500–600 s mean that IO-NCR cannot apply on them, not that the sensor biases are zeros). The gyroscope and accelerometer biases in the x-axis and the z-axis are unique and correct, but in the y-axis, the direction of rotation, IO-NCR gives the other (wrong) solution in addition to the correct one: 14.2348 deg/h for gyroscope bias and 19.5839 m/s$^2$ for accelerometer bias. It can be readily verified that the outcome is in accord with Remark 5, i.e., $|\mathbf{b}_{g+} - \mathbf{b}_{g-}| = 2\Omega \sin L \approx 14.2248$ deg/h and $|\mathbf{b}_{a+} - \mathbf{b}_{a-}| = 2g \approx 19.5834$ m/s$^2$.

Figs. 9 and 10 present EKF estimates of attitude and sensor biases, respectively. All bias estimates converge quickly to the correct values once the rotation takes place, although the gyroscope bias in the up direction needs more time to do so (see Fig. 10, top graph). It indicates that the rotation motion greatly affects observability. Fig. 11 "zooms in" on the bias estimates, where we observe an interesting phenomenon that exists in almost each run. After the strapdown INS stops at 500 s, the gyroscope estimates approach the true values faster, whereas the accelerometer estimates depart from their true values. As discussed in Remark 8, although the static state does nothing to improve observability (in addition to the rotating motion), it seemingly helps mitigate numerical errors. The reason is intuitive: it might be because various constraints have different error-propagating characteristics. From the viewpoint of global observability, the static segment in 500–600 s reinforces the constraint equations found in (9) and (10), of which (10) is used on the rotating segment in 100–500 s, as explained in the development of (23) and (24), but (9) is not used. In addition, we examine the EKF response by setting the initial value of accelerometer bias to $[0, 18 \text{ m/s}^2, 0]^T$ with other initial parameters unchanged. In Figs. 12 and 13, EKF unsurprisingly stabilizes at the unobservable state (the roll angle rests at 180 deg).

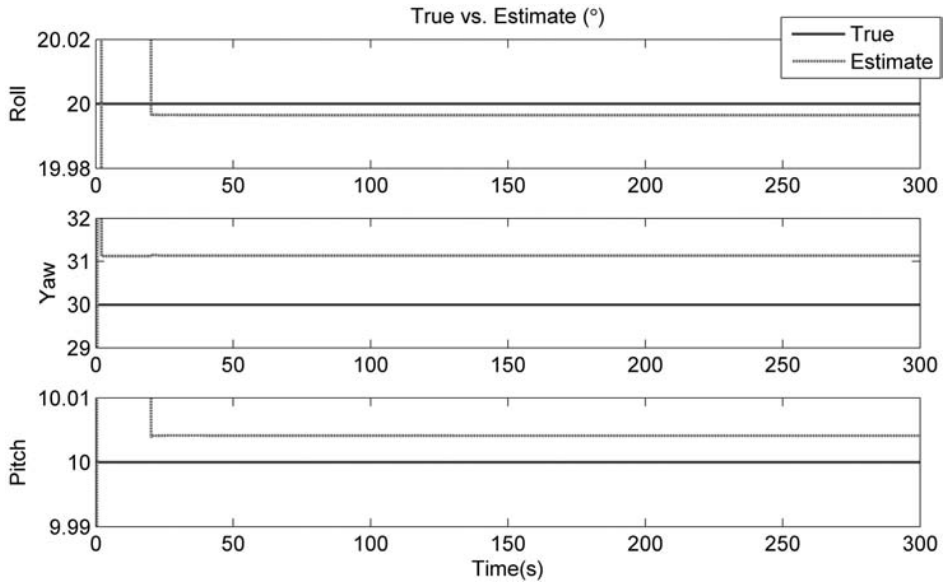

Fig. 4. Attitude estimates for other initial angle $[20.124 \text{ deg } 31.123 \text{ deg } 10.474 \text{ deg}]^T$.

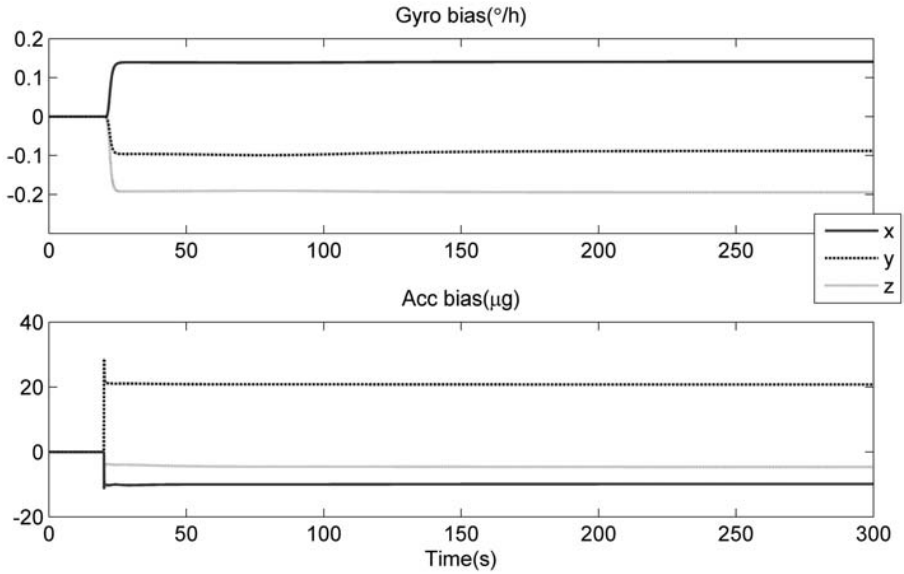

Fig. 5. Sensor bias estimates for other initial angle $[20.124 \text{ deg } 31.123 \text{ deg } 10.474 \text{ deg}]^T$.

Next, the strapdown INS rotates with the varying-magnitude $\omega_{nb}^b = [0, 6 + 4\sin(0.04\pi(t - 100)) \text{ deg/s}, 0]^T$. IO-NFVR is applied to the varying-magnitude rotation segment and yields the same result as IO-NCR in Fig. 8. Figs. 14 and 15 plot the EKF bias estimates for zero initial sensor biases and for initial accelerometer bias $[0, 18 \text{ m/s}^2, 0]^T$, respectively. EKF in this case converges to the correct or the wrong solution depending on the initial value. Because the observers perform quite similarly for constant and varying-magnitude rotations, we just present the result of the former in the next subsection.

2) *North-South Direction*: The strapdown INS is rotated along the north-south direction at $\omega_{nb}^b = [10 \text{ deg/s}, 0, 0]^T$. The IO-NCR estimates of sensor biases for 600 s are given in Fig. 16. The accelerometer biases in three axes and the gyroscope biases in the y-axis and the z-axis have only one solution, but the gyroscope bias along the rotation direction has two solutions: the correct one and 26.5164 deg/h. As revealed in Remark 5, in this case, $|\mathbf{b}_{g+} - \mathbf{b}_{g-}| = 2\Omega \cos L \approx 26.5064$ deg/h and $|\mathbf{b}_{a+} - \mathbf{b}_{a-}| = 0$. Figs. 17 and 18 present the EKF estimates of attitude and sensor biases, respectively, for 3600 s. The rotation segment is 100–3500 s. The estimates converge to the true value, although the sensor biases in the x-axis and the pitch angle exhibit apparent oscillation. EKF converges to the wrong solution for the initial gyroscope bias $[25 \text{ deg/h}, 0, 0]^T$, as shown in Figs. 19 and 20, in which the accelerometer bias in the x-axis needs more time to stabilize.

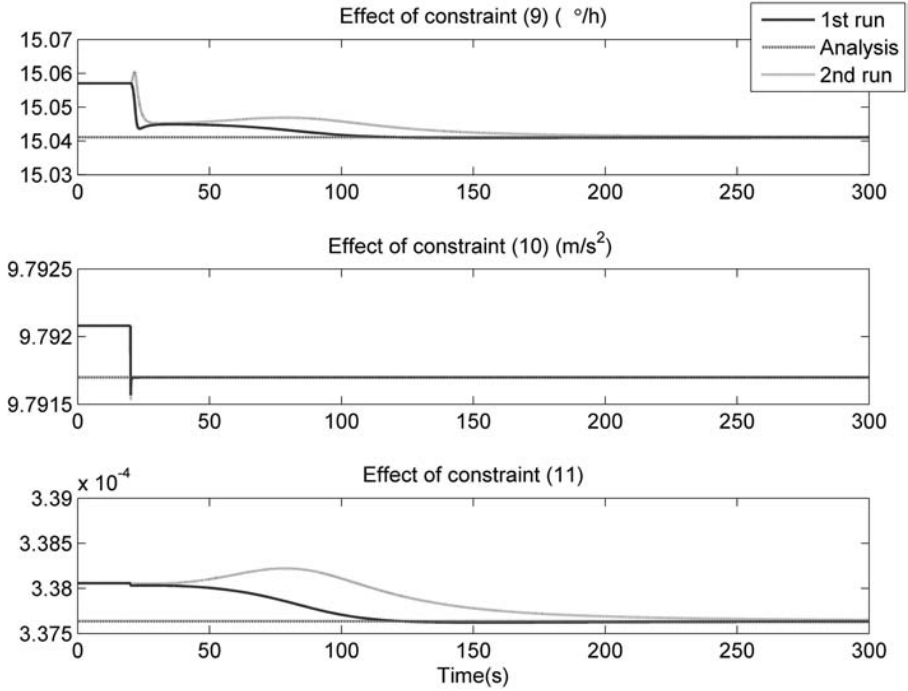

Fig. 6. Computed values of $|\omega_{ib}^b - \mathbf{b}_g|$, $|\mathbf{f}^b - \mathbf{b}_a|$, and $(\omega_{ib}^b - \mathbf{b}_g)^T(\mathbf{f}^b - \mathbf{b}_a)$ using estimated sensor biases in two sample runs, as compared to their analytical values from (9)–(11). Solid blue line is for first run, dotted blue line is for second run, and dashed red line is for analysis.

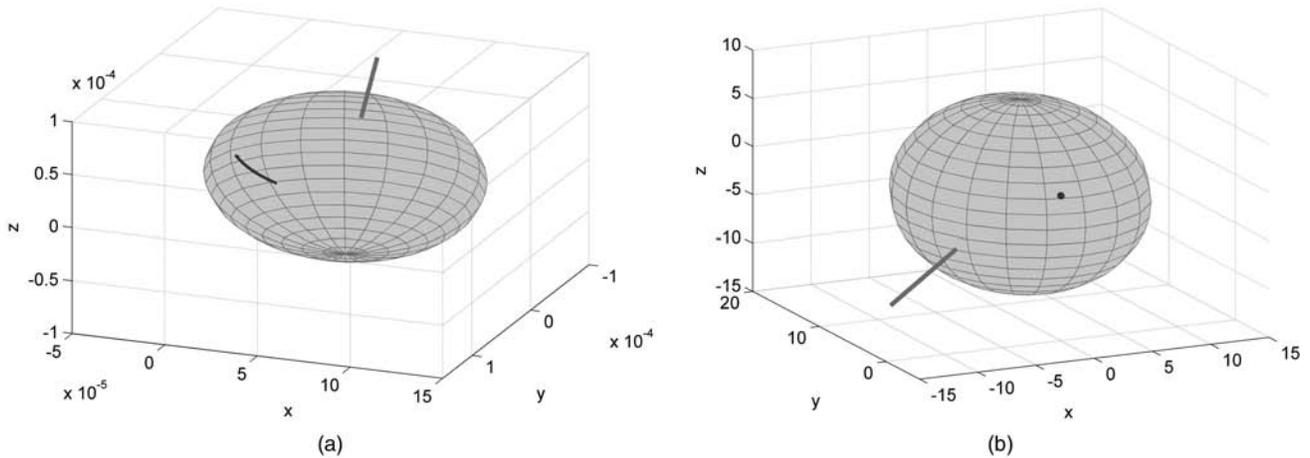

Fig. 7. Sensor bias estimates of 1000 Monte Carlo runs. Left (in radians per second): gyroscope bias estimates (blue dots), sphere surface $|\mathbf{b}_g - \omega_{ib}^b|$ (gray), and vector $-\mathbf{f}^b$ pointing outward (red). Right (in meters per second squared): accelerometer bias estimates (blue dots), sphere surface $|\mathbf{b}_a - \mathbf{s}^b|$ (gray), and vector $-\omega_{ib}^b$ pointing outward (red).

*3) East-West Direction:* The strapdown INS is rotated along the east-west direction with $\omega_{nb}^b = [0, 0, 10 \text{ deg/s}]^T$. The IO-NCR estimates of sensor biases for 600 s are given in Fig. 21. We see that the sensor biases have only one solution, because the system is observable for this case. As in Remark 5, $|\mathbf{b}_{g+} - \mathbf{b}_{g-}| = |\mathbf{b}_{a+} - \mathbf{b}_{a-}| = 0$. The EKF estimates of attitude and sensor biases for 7200 s are given in Figs. 22 and 23, respectively. The rotating period is 100–7000 s. The attitude shows severe jumps

12   IEEE TRANSACTIONS ON AEROSPACE AND ELECTRONIC SYSTEMS   VOL. 48, NO. 1   JANUARY 2012Mt2 ☐ job no. 2158 ☐ ieee ☐ aerospace and electronic systems ☐ 2158D02 [12] ☐ (XXX) ☐ 08-30-11 ☐ 03:37 PM as a result of Euler angles' ambiguity and the accelerometer bias in the z-axis converges to the true value, with strong oscillating effects. The gyroscope bias in the z-axis descends extremely slowly to 0.01 deg/h after 3000 s in Fig. 23. But we find later that the convergence speed is considerably improved for non-zero initial attitude (see Fig. 26, 100–700 s).

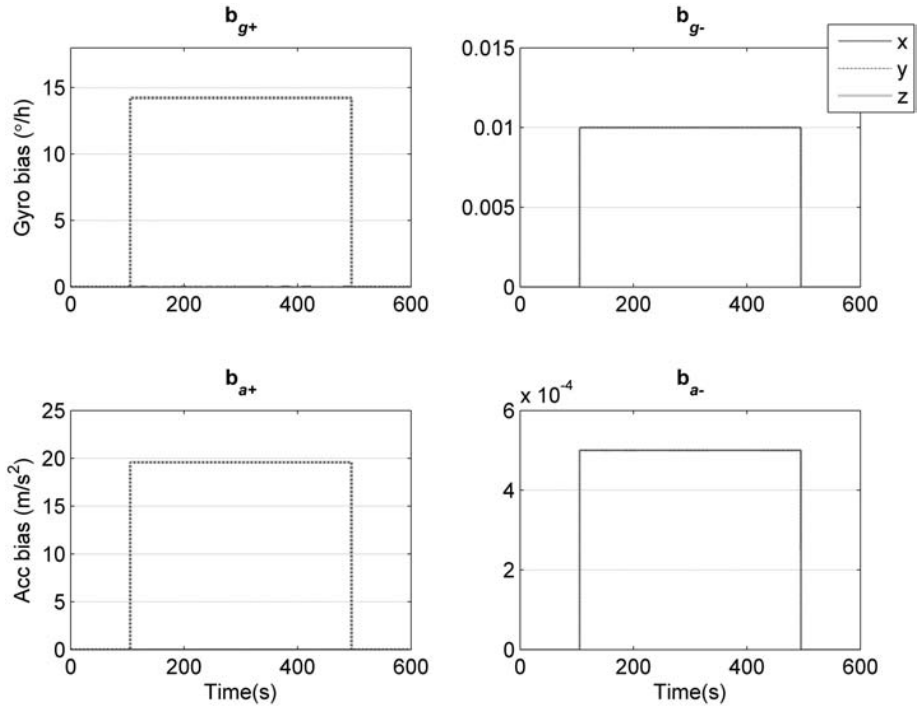

Fig. 8. Bias estimates of IO-NCR for constant up-down rotation.

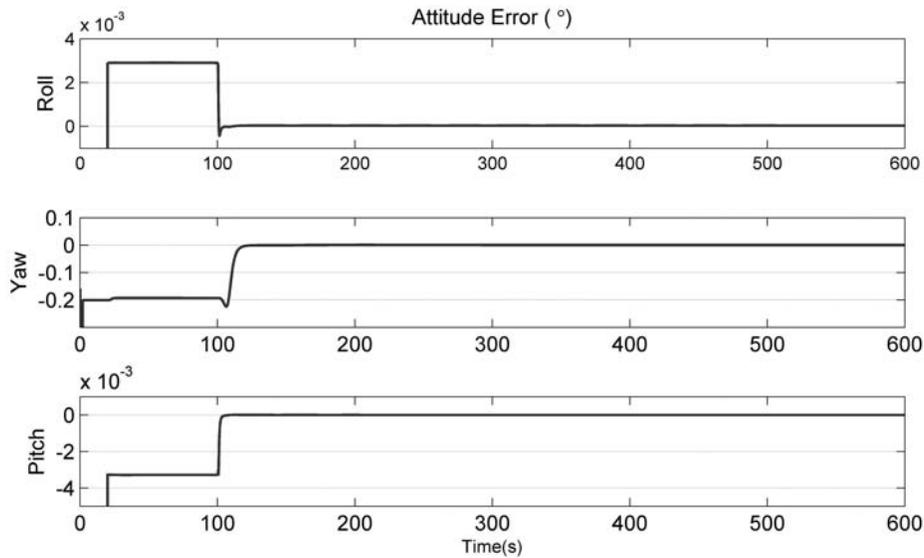

Fig. 9. Attitude estimates by EKF for constant up-down rotation.

C. Tumbling Alignment by Multiple-Axis Rotation

We consider now the tumbling alignment by rotating about multiple independent axes. The strapdown INS true attitude is set to $\varphi = 20$ deg, $\psi = 30$ deg, and $\theta = 10$ deg, and the process lasts 2400 s. The random error added to coarse alignment increases to 0.5 deg ($1\sigma$) for better demonstration. The strapdown INS rotates first in the east-west direction by $\omega_{nb}^n = [0, 0, 3 \text{ deg/s}]^T$ at 100–700 s, second in the north-south direction by $\omega_{nb}^n = [3 \text{ deg/s}, 0, 0]^T$ at 800–1400 s, and finally in the vertical direction by $\omega_{nb}^n = [0, 3 \text{ deg/s}, 0]^T$ at 1500–2100 s. For the remaining periods, the strapdown INS stays static. Fig. 24 shows the history profile for $\omega_{nb}^b$ over the three disconnected rotation segments, and Figs. 25 and 26 give the EKF estimates of attitude and sensor biases, respectively. The estimates converge satisfactorily to their true values. The second rotation starting at 800 s removes the remaining solution ambiguity after the first rotation[4] and has a great positive impact on reducing the bias estimate errors and attitude errors, especially the accelerometer biases in the y- and z-axes. The EKF computed standard variances of bias estimates are plotted in Fig. 27.

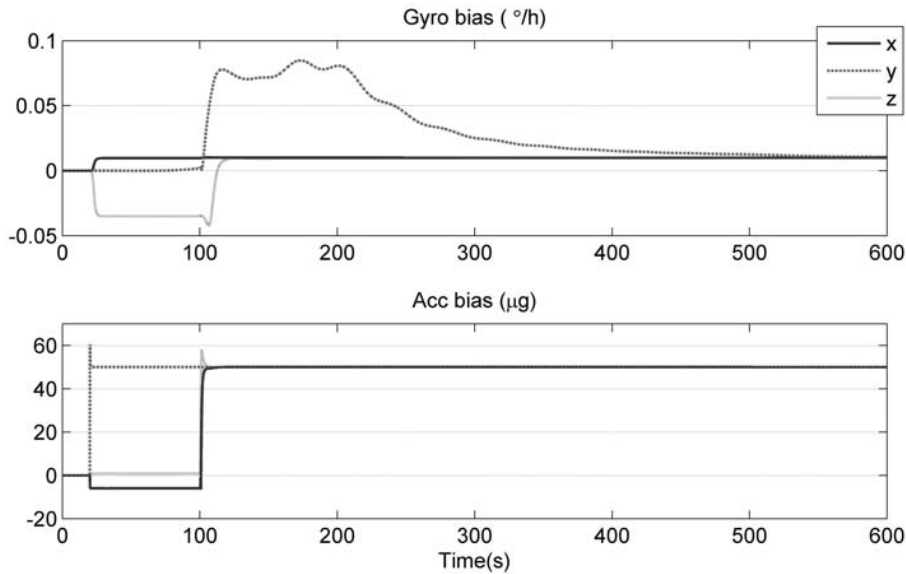

Fig. 10. Sensor bias estimates by EKF for constant up-down rotation.

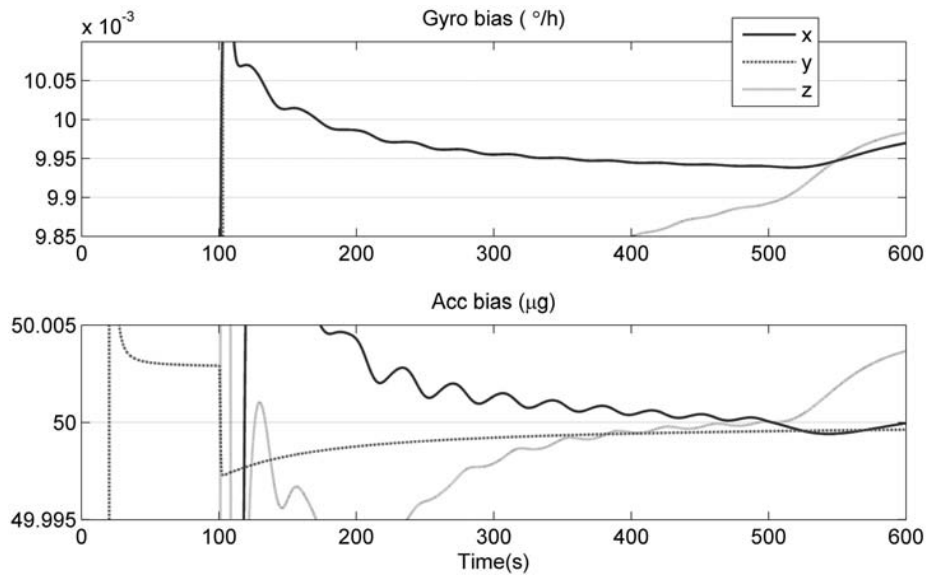

Fig. 11. "Zoom-in" view of EKF bias estimates.

Finally, we report an interesting phenomenon about the wrong solution. The true initial angles are set to zeros for easy demonstration. The strapdown INS rotates first in the up-down direction by $\omega_{nb}^n = [0, 10 \deg/s, 0]^T$ at 100–500 s and second in the north-south direction by $\omega_{nb}^n = [10 \deg/s, 0, 0]^T$ at 600–1000 s. The first rotation is the same as in the single up-down rotation in subsection B. Figs. 28 and 29 present the EKF results when setting the initial value of accelerometer bias to $[0, 18 \text{ m/s}^2, 0]^T$. EKF stabilizes at the unobservable and wrong solution during the first up-down rotation but is forced to diverge from the wrong solution once the second rotation starts. The second rotation eliminates the solution ambiguity, and the wrong solution is no longer a stabilizing state for EKF. This phenomenon is a convincing support for the analysis in Section III.

## V. CONCLUSIONS AND FUTURE WORKS

In this paper, we have revisited the strapdown INS static and tumbling alignment from the perspective of global observability. The observability problem is formulated as SAOP, i.e., whether it suffices to determine the initial state by solving the infinite nonlinear equations over the continuous time interval. Equivalently, it investigates the effect of the known body angular rate and specific force on state observability. We prove that it is not the static positions but the rotating motion that matters for observability. Furthermore, SAOP will be fully observable if the strapdown INS is rotated successively about two axes. For cases of rotating about a single axis, SAOP will be nearly observable for no more than two unobservable states to which the explicit solutions are analytically derived. The global observability analysis is shown to be straightforward and constructive and results in insights into and a clearer picture of the strapdown INS alignment problem. It sheds light on the incompleteness and inconsistency of previous results. The paper also throws doubts on and calls for a review of all linearization-based observability studies in the vast literature.

---

[4]Thanks to one reviewer's suggestions, it can be shown that an added linear acceleration helps resolve the ambiguity. See Appendix Section E for a brief discussion.

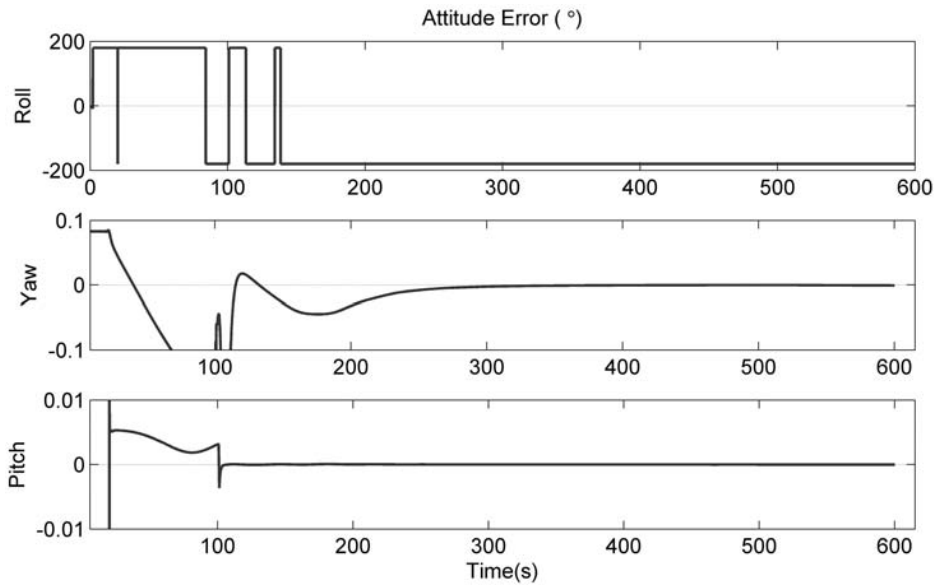

Fig. 12. Attitude estimates by EKF (wrong convergence) for constant up-down rotation. Apparent jumps in top graph result from Euler angles' ambiguity.

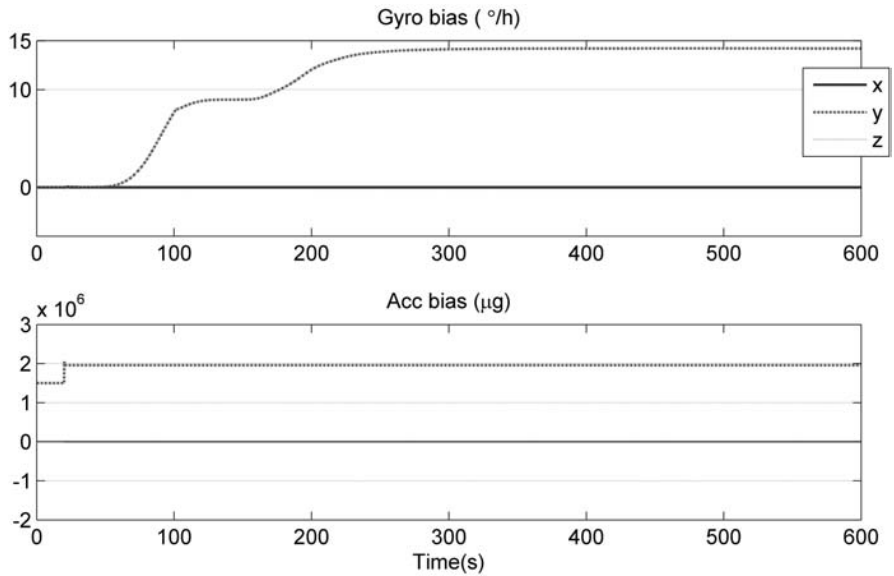

Fig. 13. Sensor bias estimates by EKF (wrong convergence) for constant up-down rotation.

The theorems and claims in this paper are supported by extensive simulations with constructed ideal observers and an approximate nonlinear observer (EKF). Although they make no observability contribution in theory, EKF results show that static segments do help mitigate numerical errors in the rotating segments and static segments interlaced simulations. Conclusions obtained can assist in optimal tumbling strategy and appropriate state observer designs in practice to improve the alignment performance. Before that, however, we have to consider the lever arm between the table rotation center and the strapdown INS origin, because presence of the lever arm may remarkably decay the accuracy of the zero velocity measurement and thus the estimation performance. We are working on this in the context of global observability.

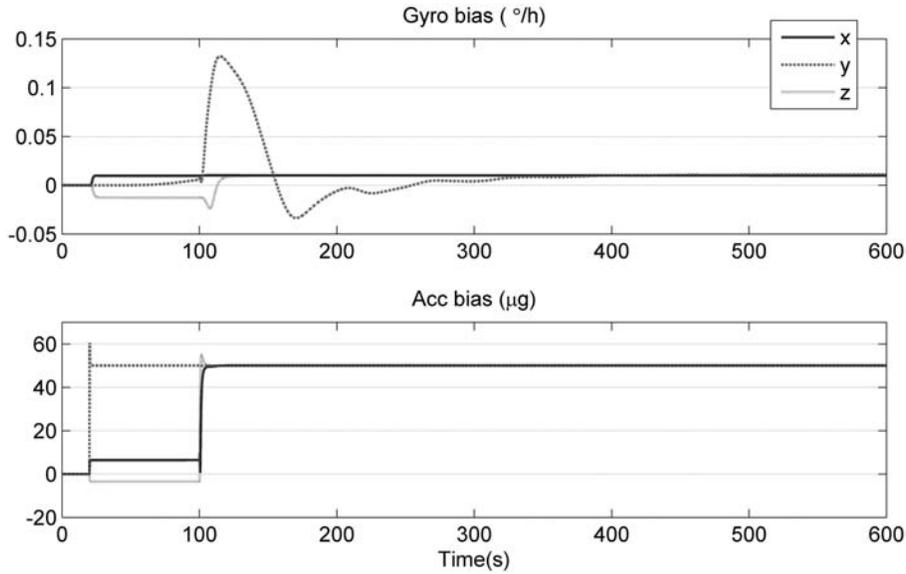

Fig. 14. Sensor bias estimates by EKF for varying-magnitude up-down rotation.

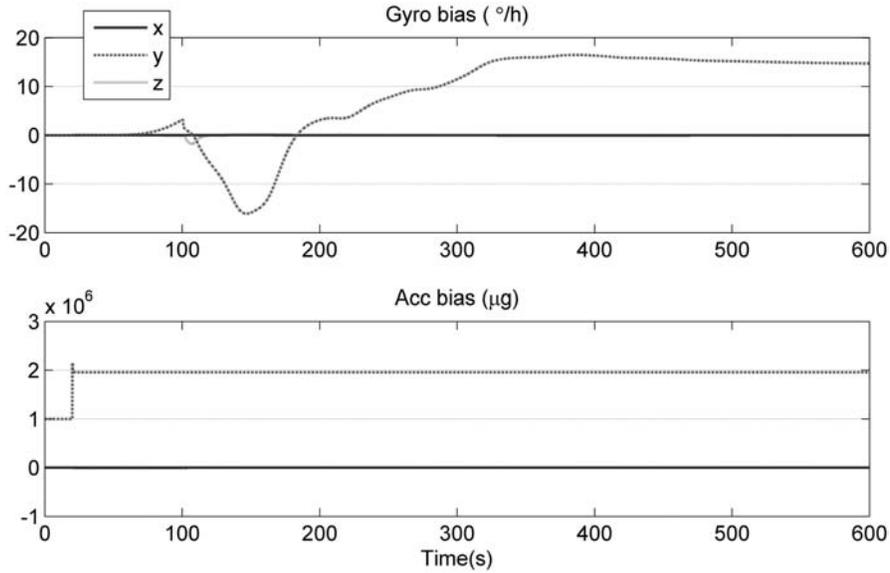

Fig. 15. Sensor bias estimates by EKF (wrong convergence) for varying-magnitude up-down rotation.

ACKNOWLEDGMENT

Thanks to Ph.D. student Dayong Zhang for partial simulation assistance. The authors acknowledge Reviewer 2 for many detailed suggestions that considerably improved the content and presentation of early drafts of the paper.

APPENDIX

A. Proof of Lemma 2

The equality $|\mathbf{a}_k - \mathbf{x}| = r$ means that $\mathbf{x}$ locates on the surface of the sphere with radius $r$ that centers on $\mathbf{a}_k$, i.e.,

$$\mathbf{a}_k^T \mathbf{a}_k - 2\mathbf{a}_k^T \mathbf{x} + \mathbf{x}^T \mathbf{x} = r^2, \qquad k = 1, 2, \ldots, m. \quad (49)$$

Without loss of generality, subtracting (49) for $k = 1$ from (49) for $k = 2, \ldots, m$ yields, in the matrix form,

$$2 \begin{bmatrix} (\mathbf{a}_2 - \mathbf{a}_1)^T \\ \vdots \\ (\mathbf{a}_m - \mathbf{a}_1)^T \end{bmatrix} \mathbf{x} = \begin{bmatrix} |\mathbf{a}_2|^2 - |\mathbf{a}_1|^2 \\ \vdots \\ |\mathbf{a}_m|^2 - |\mathbf{a}_1|^2 \end{bmatrix} \quad (50)$$

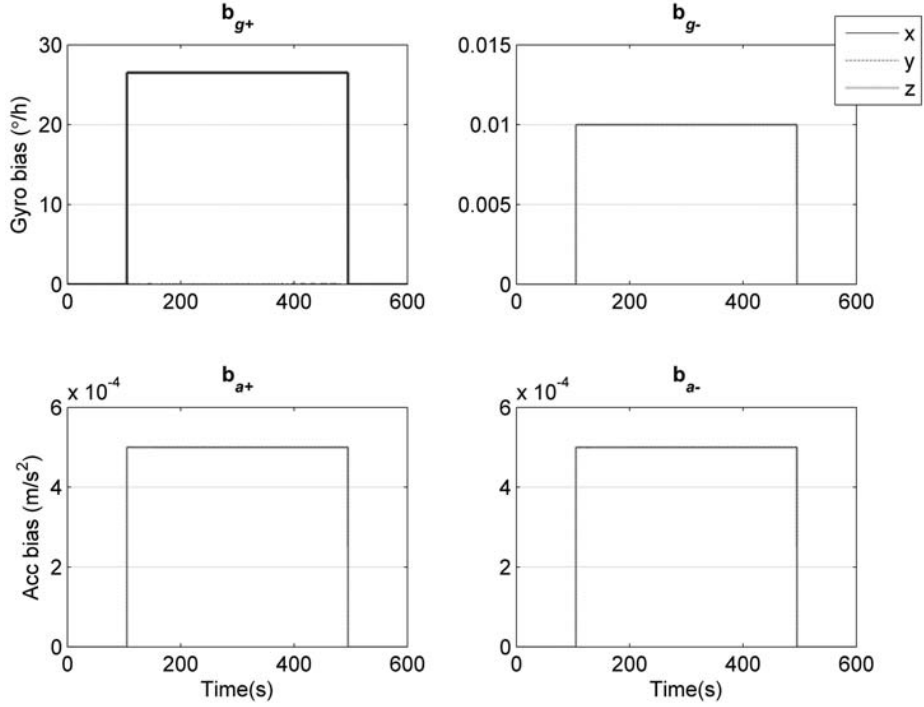

Fig. 16. Bias estimates of IO-NCR for constant north-south rotation.

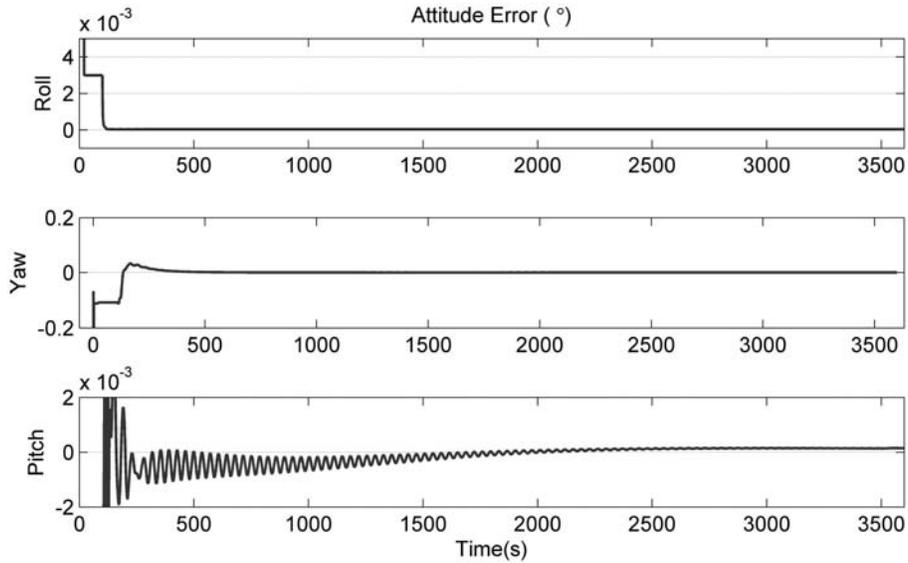

Fig. 17. Attitude estimates of EKF for constant north-south rotation.

each row of which represents a plane in geometry. Left multiplying $[\mathbf{a}_2 - \mathbf{a}_1 \cdots \mathbf{a}_m - \mathbf{a}_1]$ on both sides, we have

$$2\left(\sum_{k=2}^{m}(\mathbf{a}_k - \mathbf{a}_1)(\mathbf{a}_k - \mathbf{a}_1)^T\right)x$$
$$= \sum_{k=2}^{m}(\mathbf{a}_k - \mathbf{a}_1)(|\mathbf{a}_k|^2 - |\mathbf{a}_1|^2). \quad (51)$$

If the matrix $\mathbf{A} \stackrel{\Delta}{=} \sum_{k=2}^{m}(\mathbf{a}_k - \mathbf{a}_1)(\mathbf{a}_k - \mathbf{a}_1)^T$ is nonsingular or, equivalently, points $\mathbf{a}_k$ do not lie in one plane, the unknown $\mathbf{x}$ can be determined. Nonsingular $\mathbf{A}$ implies $m \geq 4$, because any three points are contained in some common plane.

For singular cases, we have further comments:

1) If rank($\mathbf{A}$) = 2, i.e., $\mathbf{a}_k$ lie in a common plane (but not on a line), then the solution space of (50) is of dimension 1. For any feasible $\mathbf{x}$, the line $\mathbf{x} + \alpha \zeta$ also satisfies (50), in which $\alpha$ is a real scalar and $\zeta$ is the unit normal of the plane. Therefore, we may get one or two solutions, depending on the relationship between the line $\mathbf{x} + \alpha \zeta$ and the sphere surface, i.e., (49) for $k = 1$.

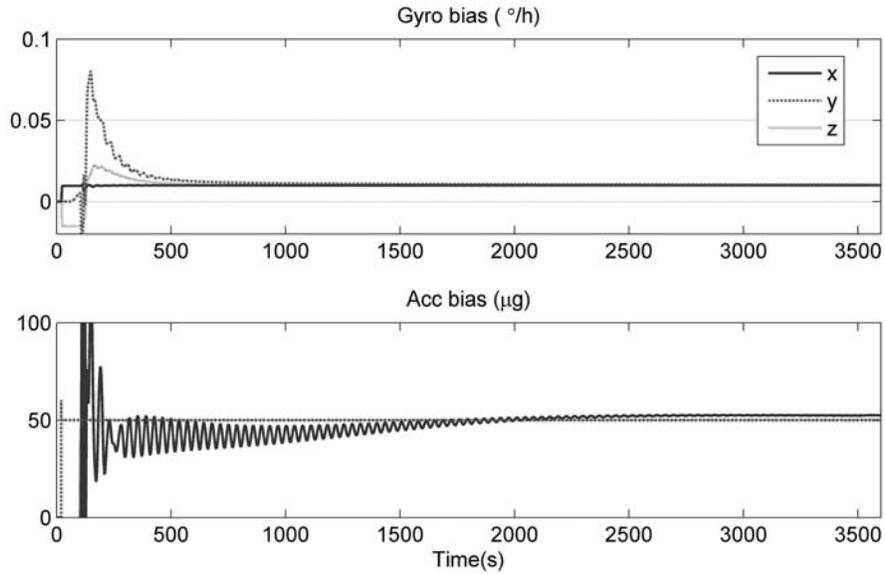

Fig. 18. Sensor bias estimates by EKF for constant north-south rotation.

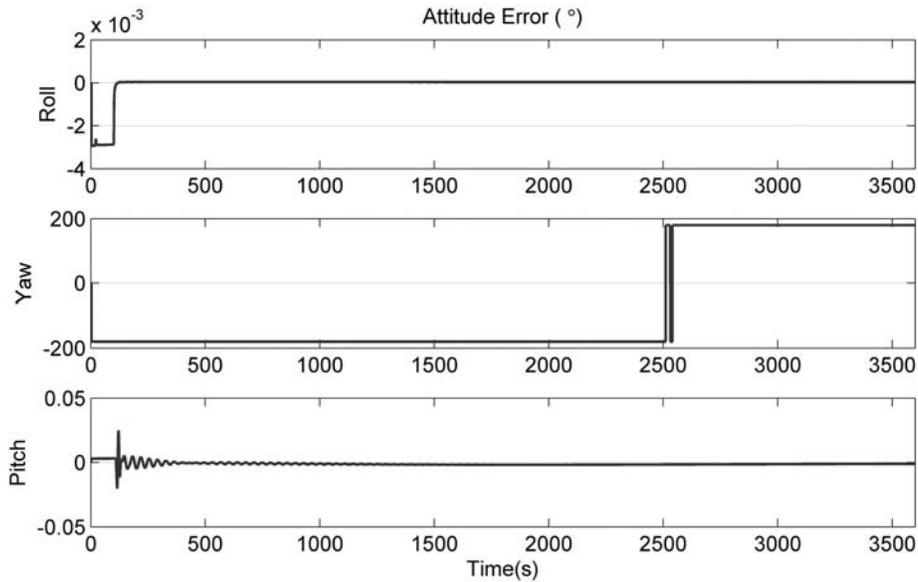

Fig. 19. Attitude estimates of EKF (wrong solution) for constant north-south rotation. Apparent jumps in middle graph result from Euler angles' singularity.

2) If rank(**A**) = 1, i.e., $\mathbf{a}_k$ lie on a common line—say, **l**—then the solution space is of dimension 2. The possible solutions lie on the intersection of the sphere surface, i.e., (49) for $k = 1$, and the plane normal to **l**.

3) If rank(**A**) = 0, i.e., $\mathbf{a}_k$ are equal to each other. The solution space is the sphere surface (49) for $k = 1$.

B. Proof of Lemma 3

Because $\mathbf{a}(t)$ has nonconstant directions in the interval, there exist two time instants—say, $t_1$ and $t_2$—in the interval such that $\mathbf{a}(t_1)$ and $\mathbf{a}(t_2)$ have different directions. Then we have

$$\begin{bmatrix} \mathbf{a}(t_1)\times \\ \mathbf{a}(t_2)\times \end{bmatrix} \mathbf{m} = \begin{bmatrix} \mathbf{b}(t_1) \\ \mathbf{b}(t_2) \end{bmatrix}. \quad (52)$$

Because

$$\operatorname{rank}\left(\begin{bmatrix} \mathbf{a}(t_1)\times \\ \mathbf{a}(t_2)\times \end{bmatrix}\right) = 3$$

**m** has a unique solution.

C. Proof of Lemma 4

Because

$$\mathbf{a} \times \mathbf{m} = \mathbf{b} \quad (53)$$

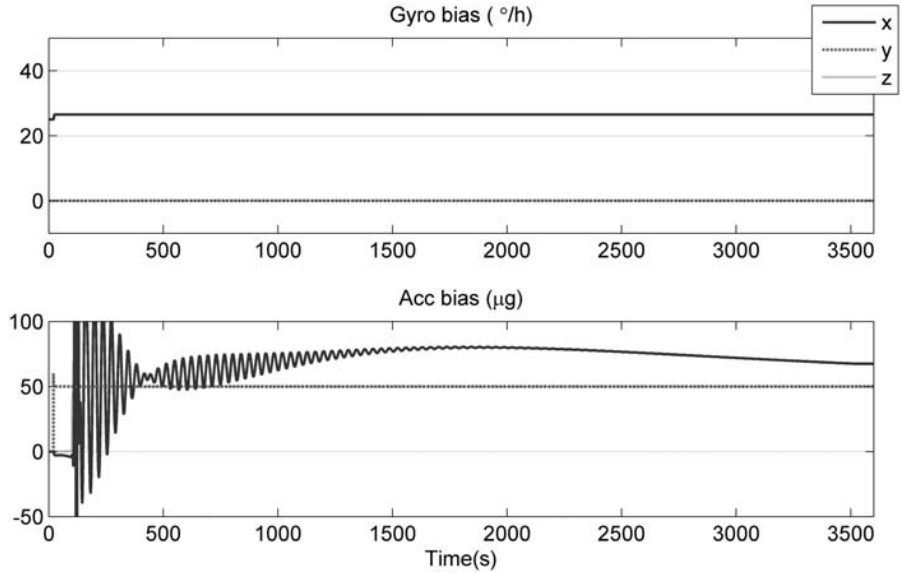

Fig. 20. Sensor bias estimates by EKF (wrong solution) for constant north-south rotation.

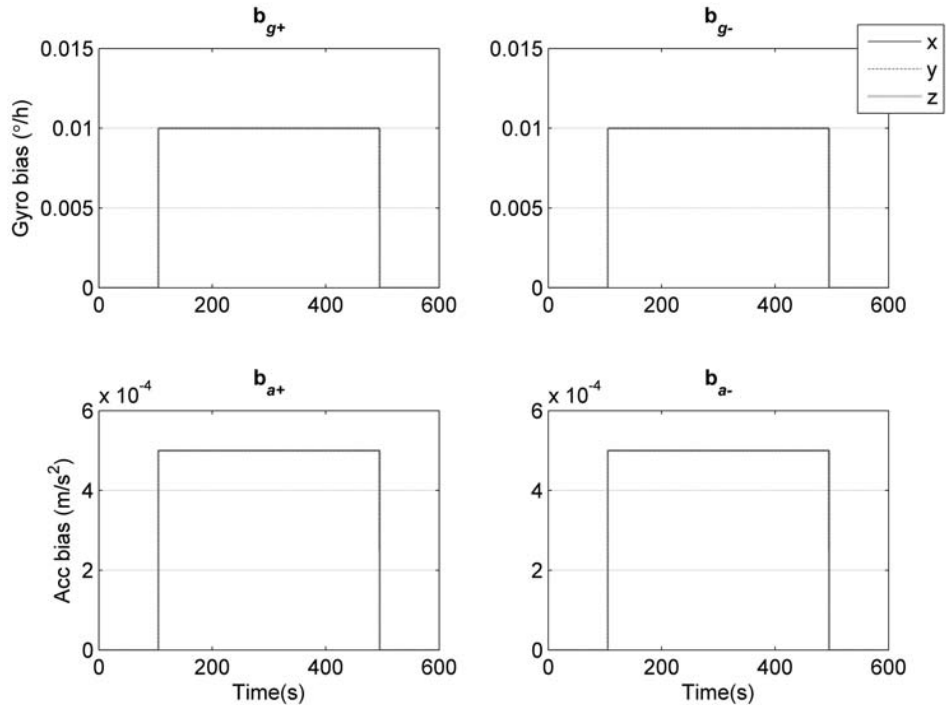

Fig. 21. Bias estimates of IO-NCR for constant east-west rotation.

the unknown vector $\mathbf{m}$ lies in the plane containing the vectors $\mathbf{a}$ and $\mathbf{a} \times \mathbf{b}$ and can be expressed as

$$\mathbf{m} = \alpha \mathbf{a} + \beta \mathbf{a} \times \mathbf{b} \qquad (54)$$

where $\alpha$ and $\beta$ are real numbers. Substituting into (53), and considering $\mathbf{a}^T \mathbf{b} = 0$,

$$\beta \mathbf{a} \times (\mathbf{a} \times \mathbf{b}) = \beta(\mathbf{a}^T \mathbf{b} \mathbf{a} - |\mathbf{a}|^2 \mathbf{b}) = -\beta |\mathbf{a}|^2 \mathbf{b} = \mathbf{b}. \qquad (55)$$

So $\beta = -1/|\mathbf{a}|^2$. Taking the norm on both sides, (54) gives

$$|\mathbf{m}|^2 = \alpha^2 |\mathbf{a}|^2 + \beta^2 |\mathbf{a} \times \mathbf{b}|^2 = \alpha^2 |\mathbf{a}|^2 + \beta^2 |\mathbf{a}|^2 |\mathbf{b}|^2 \qquad (56)$$

which yields $\alpha = \pm\sqrt{|\mathbf{a}|^2 |\mathbf{m}|^2 - |\mathbf{b}|^2}/|\mathbf{a}|^2$. The solution of $\mathbf{m}$ is

$$\mathbf{m} = \pm \frac{\mathbf{a}\sqrt{|\mathbf{a}|^2|\mathbf{m}|^2 - |\mathbf{b}|^2}}{|\mathbf{a}|^2} - \frac{\mathbf{a} \times \mathbf{b}}{|\mathbf{a}|^2} = \pm \frac{\mathbf{a}|\mathbf{a} \cdot \mathbf{m}|}{|\mathbf{a}|^2} - \frac{\mathbf{a} \times \mathbf{b}}{|\mathbf{a}|^2}. \qquad (57)$$

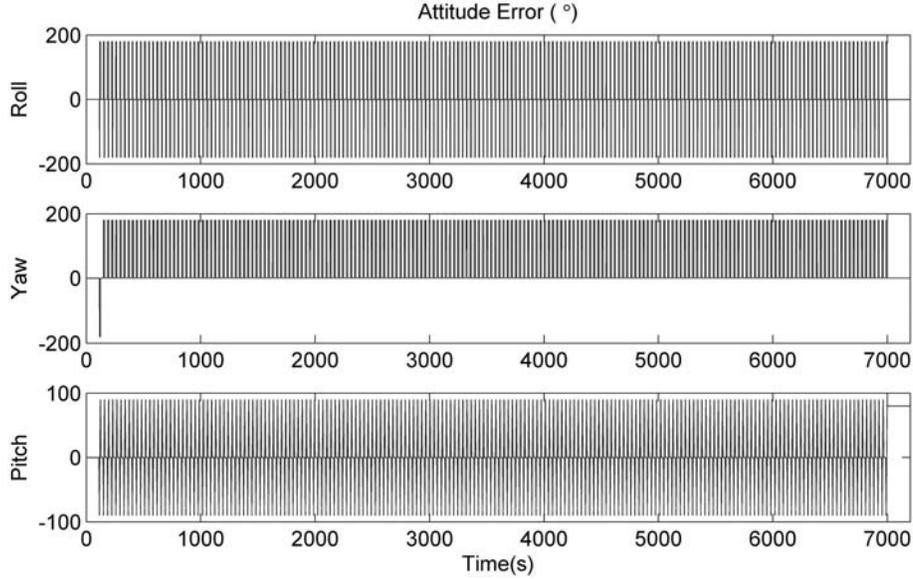

Fig. 22. Attitude estimates of EKF for constant east-west rotation. Apparent jumps result from Euler angles' ambiguity.

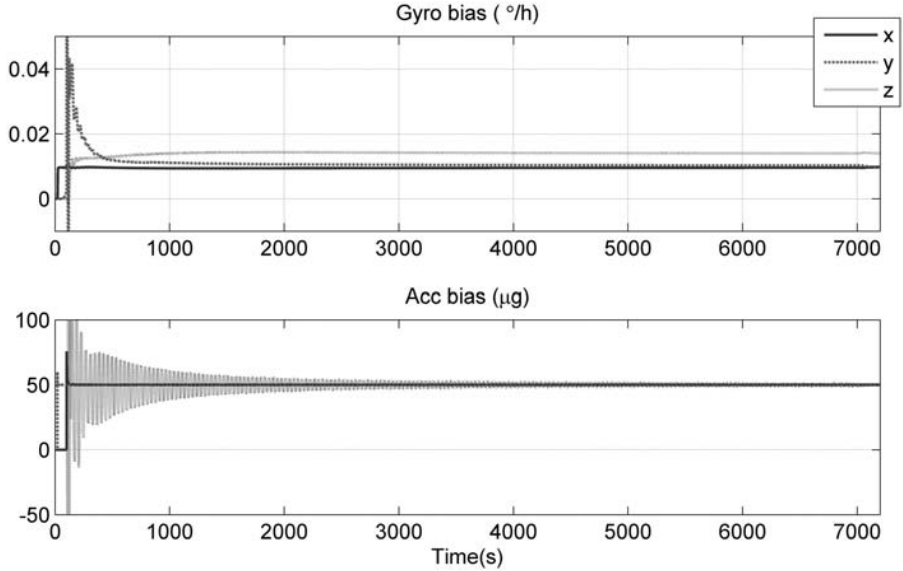

Fig. 23. Sensor bias estimates by EKF for constant east-west rotation.

D. Why Are Only Two of Four Pairs Valid?

Check the angular relationship of $\boldsymbol{\omega}_{ie}^b$ and $\mathbf{g}^b$, which is similar to (11). Substituting (25) and (28),

$$-\boldsymbol{\omega}_{ie}^b \cdot \mathbf{g}^b = (\boldsymbol{\omega}_{ib}^b - \mathbf{b}_{g+,-} - \boldsymbol{\omega}_{nb}^b) \cdot (\mathbf{f}^b - \mathbf{b}_{a+,-})$$

$$= \left( \mp \frac{\boldsymbol{\omega}_{nb}^b |\boldsymbol{\omega}_{nb}^b \cdot \boldsymbol{\omega}_{ie}^b|}{|\boldsymbol{\omega}_{nb}^b|^2} + \frac{\boldsymbol{\omega}_{nb}^b \times \dot{\boldsymbol{\omega}}_{ib}^b}{|\boldsymbol{\omega}_{nb}^b|^2} \right)$$

$$\cdot \left( \mp \frac{\boldsymbol{\omega}_{nb}^b |\boldsymbol{\omega}_{nb}^b \cdot \mathbf{g}^b|}{|\boldsymbol{\omega}_{nb}^b|^2} + \frac{\boldsymbol{\omega}_{nb}^b \times \dot{\mathbf{f}}^b}{|\boldsymbol{\omega}_{nb}^b|^2} \right)$$

$$= \pm \frac{|\boldsymbol{\omega}_{nb}^b \cdot \boldsymbol{\omega}_{ie}^b||\boldsymbol{\omega}_{nb}^b \cdot \mathbf{g}^b|}{|\boldsymbol{\omega}_{nb}^b|^2} + \frac{(\boldsymbol{\omega}_{nb}^b \times \dot{\boldsymbol{\omega}}_{ib}^b) \cdot (\boldsymbol{\omega}_{nb}^b \times \dot{\mathbf{f}}^b)}{|\boldsymbol{\omega}_{nb}^b|^4}.$$

(58)

For the third equality, "+" indicates $\mathbf{b}_a$ and $\mathbf{b}_g$ take the same sign; otherwise, they take different signs. From (23) and (26), $\boldsymbol{\omega}_{nb}^b$ is perpendicular to both $\dot{\boldsymbol{\omega}}_{ib}^b$ and $\dot{\mathbf{f}}^b$ and

$$-\boldsymbol{\omega}_{ie}^b \cdot \mathbf{g}^b = \pm \frac{|\boldsymbol{\omega}_{nb}^b \cdot \boldsymbol{\omega}_{ie}^b||\boldsymbol{\omega}_{nb}^b \cdot \mathbf{g}^b|}{|\boldsymbol{\omega}_{nb}^b|^2} + \frac{\dot{\boldsymbol{\omega}}_{ib}^b \cdot \dot{\mathbf{f}}^b}{|\boldsymbol{\omega}_{nb}^b|^2}$$

$$= \pm \frac{|\boldsymbol{\omega}_{nb}^b \cdot \boldsymbol{\omega}_{ie}^b||\boldsymbol{\omega}_{nb}^b \cdot \mathbf{g}^b|}{|\boldsymbol{\omega}_{nb}^b|^2} - \frac{(\boldsymbol{\omega}_{nb}^b \times \mathbf{g}^b) \cdot (\boldsymbol{\omega}_{nb}^b \times \boldsymbol{\omega}_{ie}^b)}{|\boldsymbol{\omega}_{nb}^b|^2}$$

$$= \pm \frac{|(\boldsymbol{\omega}_{nb}^b \cdot \boldsymbol{\omega}_{ie}^b)(\boldsymbol{\omega}_{nb}^b \cdot \mathbf{g}^b)|}{|\boldsymbol{\omega}_{nb}^b|^2} + \frac{(\boldsymbol{\omega}_{nb}^b \cdot \boldsymbol{\omega}_{ie}^b)(\boldsymbol{\omega}_{nb}^b \cdot \mathbf{g}^b)}{|\boldsymbol{\omega}_{nb}^b|^2}$$

$$-\boldsymbol{\omega}_{ie}^b \cdot \mathbf{g}^b \qquad (59)$$

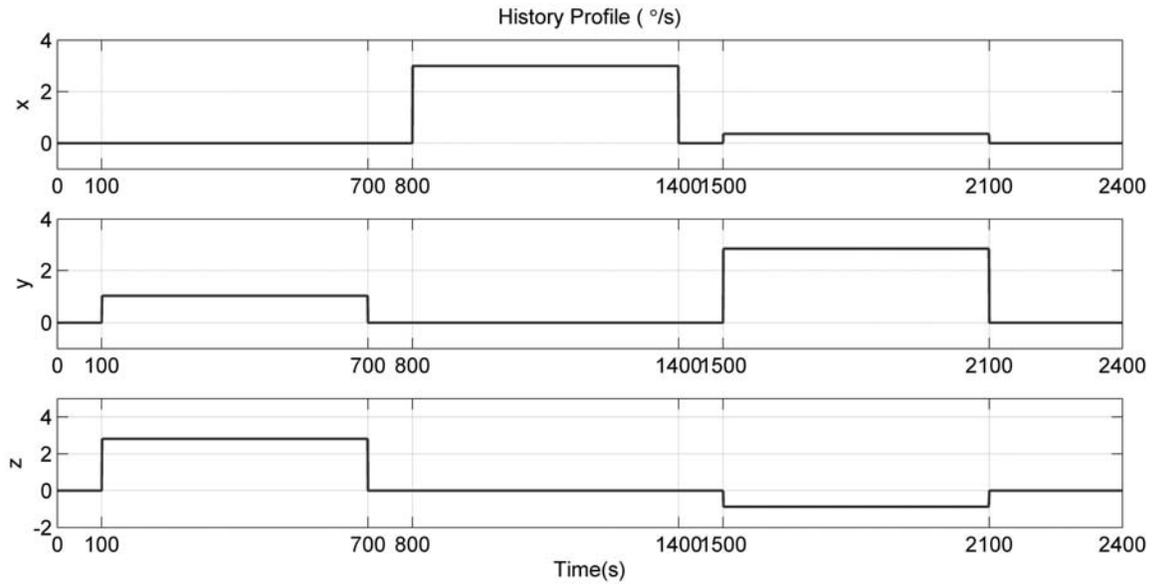

Fig. 24. History profile of $\omega_{nb}^b$ in multiple-axis tumbling alignment.

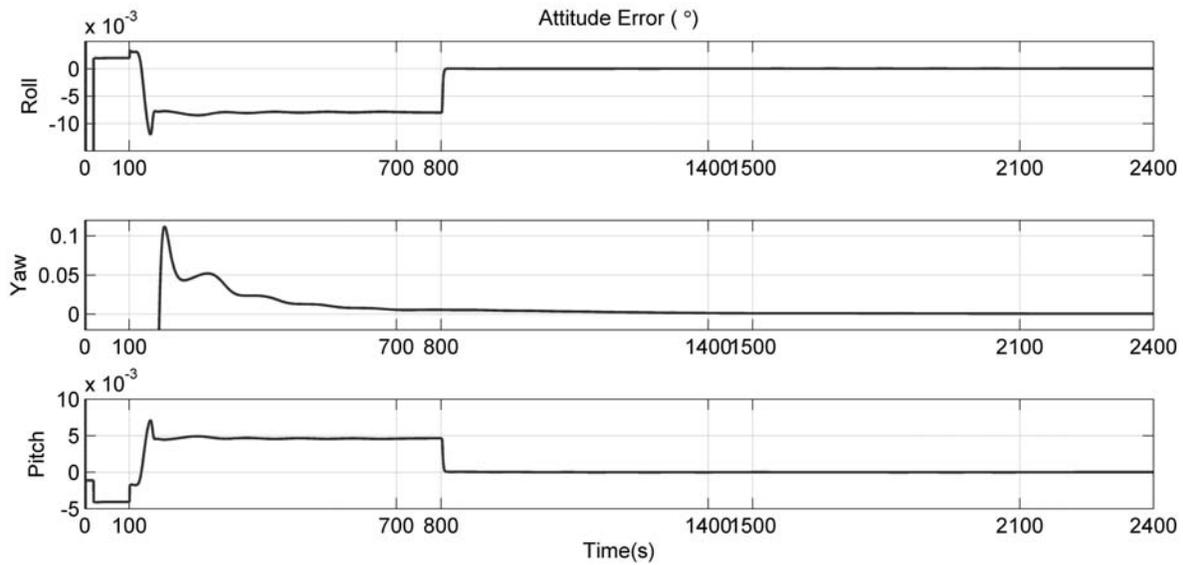

Fig. 25. EKF attitude estimates for multiple-axis tumbling alignment.

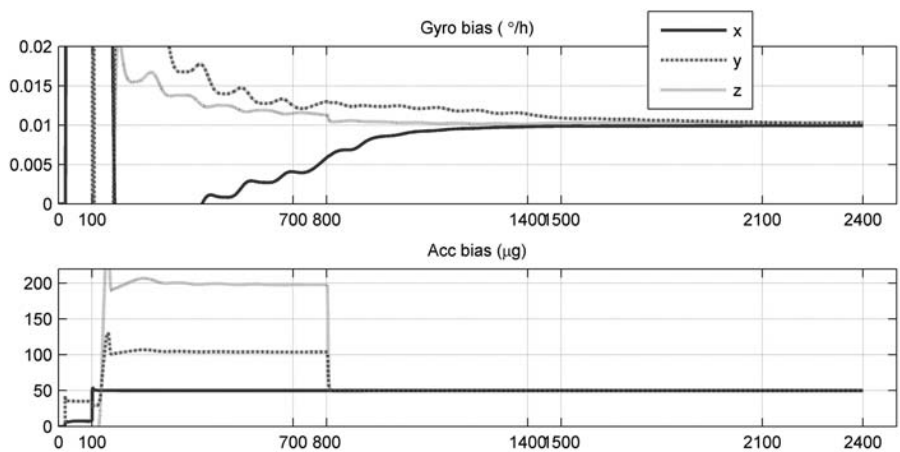

Fig. 26. EKF sensor bias estimates for multiple-axis tumbling alignment.

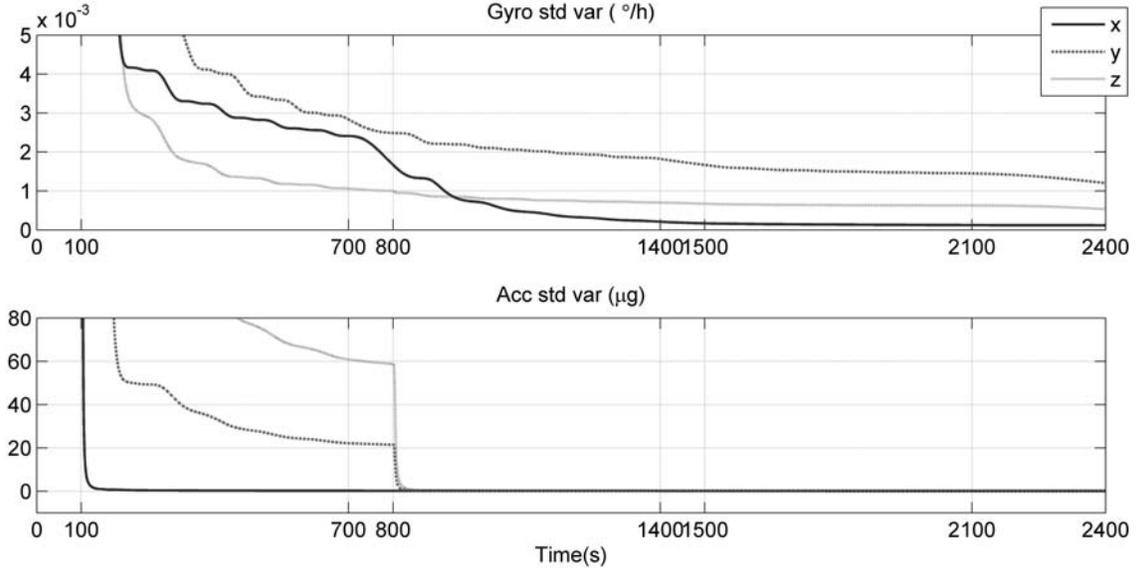

Fig. 27. EKF computed standard variances of bias estimates in multiple-axis tumbling alignment.

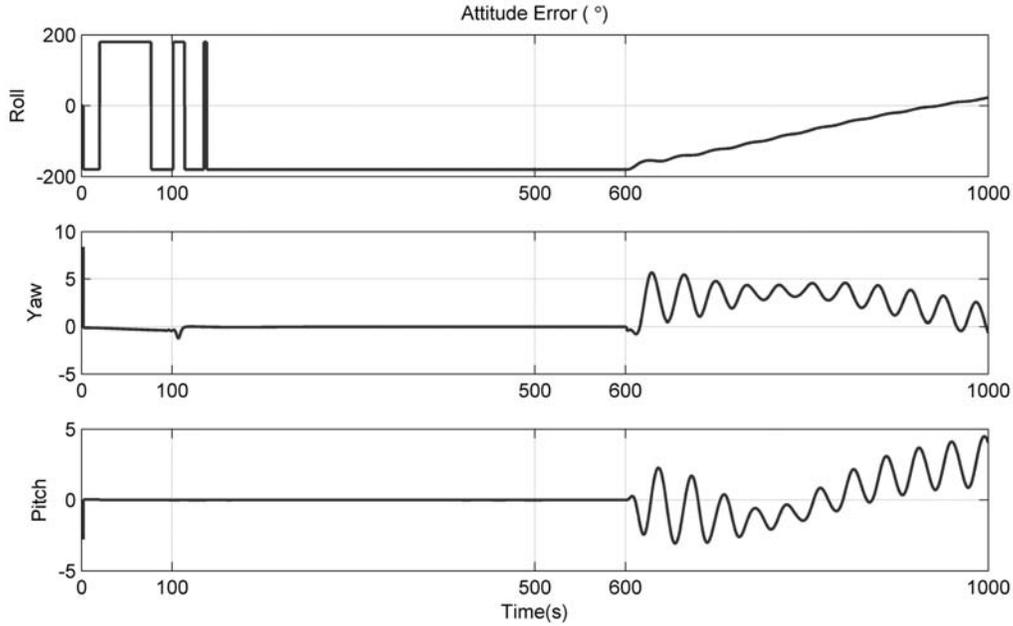

Fig. 28. EKF attitude estimate for multiple-axis tumbling alignment (wrong solution phenomenon).

which means that the equality is valid only when it takes "+" or "−," not both. If $(\boldsymbol{\omega}_{nb}^b \cdot \boldsymbol{\omega}_{ie}^b)(\boldsymbol{\omega}_{nb}^b \cdot \mathbf{g}^b) < 0$, it takes "+" and the $(\mathbf{b}_a, \mathbf{b}_g)$ pairs with same signs are feasible, i.e., $(\mathbf{b}_{a+}, \mathbf{b}_{g+})$ and $(\mathbf{b}_{a-}, \mathbf{b}_{g-})$; otherwise, $(\mathbf{b}_{a+}, \mathbf{b}_{g-})$ and $(\mathbf{b}_{a-}, \mathbf{b}_{g+})$ are feasible pairs.

### E. Added Linear Acceleration Helps Resolve the Ambiguity

Consider an added linear acceleration maneuver after the first rotation. If the non-zero reference velocity $\mathbf{v}^n$ is available while accelerated, we have from (2)

$$\dot{\mathbf{v}}^n = \mathbf{C}_b^n(\mathbf{f}^b - \mathbf{b}_a) - (2\boldsymbol{\omega}_{ie}^n + \boldsymbol{\omega}_{en}^n) \times \mathbf{v}^n + \mathbf{g}^n$$

$$\leftrightarrow \quad |\mathbf{f}^b - \mathbf{b}_a| = |\dot{\mathbf{v}}^n + (2\boldsymbol{\omega}_{ie}^n + \boldsymbol{\omega}_{en}^n) \times \mathbf{v}^n - \mathbf{g}^n|$$

$$\leftrightarrow \quad \mathbf{f}^{bT}\mathbf{f}^b - 2\mathbf{f}^{bT}\mathbf{b}_a + \mathbf{b}_a^T\mathbf{b}_a$$

$$= |\dot{\mathbf{v}}^n + (2\boldsymbol{\omega}_{ie}^n + \boldsymbol{\omega}_{en}^n) \times \mathbf{v}^n - \mathbf{g}^n|^2 \overset{\Delta}{=} \rho. \quad (60)$$

Taking the time derivative on both sides,

$$2\dot{\mathbf{f}}^{bT}\mathbf{b}_a = 2\dot{\mathbf{f}}^{bT}\mathbf{f}^b - \dot{\rho}. \quad (61)$$

Because we have from (23) $\boldsymbol{\omega}_{nb}^b \times \mathbf{b}_a = \boldsymbol{\omega}_{nb}^b \times \mathbf{f}^b + \dot{\mathbf{f}}^b$, the accelerometer bias $\mathbf{b}_a$ is obtained if $\dot{\mathbf{f}}^b$ is not parallel to $\boldsymbol{\omega}_{nb}^b$ (Lemma 3). Consequently, the gyroscope bias ambiguity is removed.

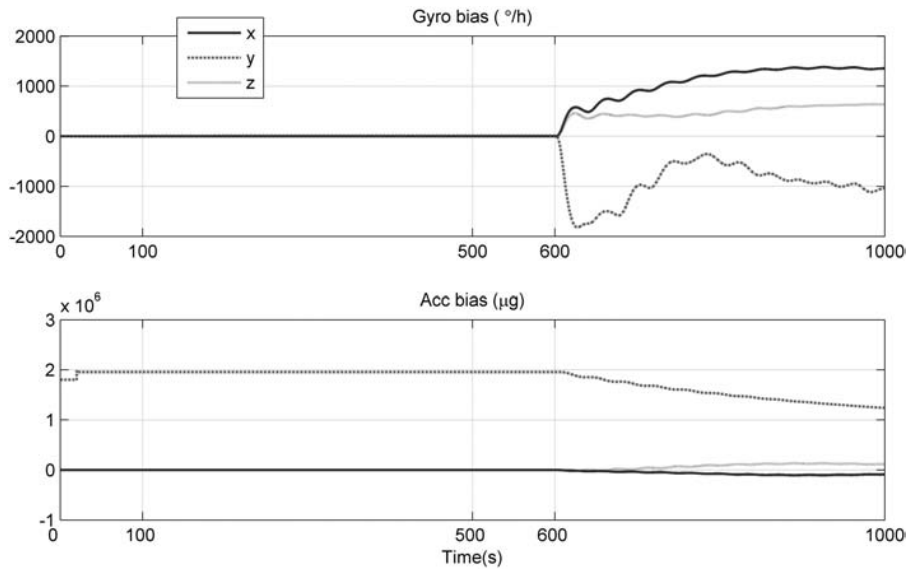

Fig. 29. EKF sensor bias estimates for multiple-axis tumbling alignment (wrong solution phenomenon).

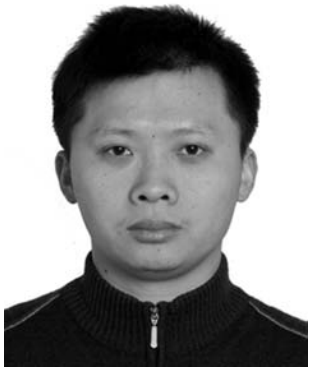

**Yuanxin Wu** was born in Jinan, People's Republic of China, in 1976. He received the B.Sc. and Ph.D. degrees in navigation from the Department of Automatic Control, National University of Defense Technology, Changsha, People's Republic of China, in 1998 and 2005, respectively.

From 2005 to 2007, he was with the National University of Defense Technology as a Lecturer; he is now an associate professor with the same university. From February 2009 to February 2010, he was a visiting Post Doctor Fellow in the Department of Geomatics Engineering, University of Calgary, Canada. His current research interests include inertial navigation systems, inertial-based integrated navigation systems, and state estimation theory.

Dr. Wu was the recipient of 2008 Top 100 National Excellent Doctoral Dissertations in China and 2010 New Century Excellent Talents in University in China.

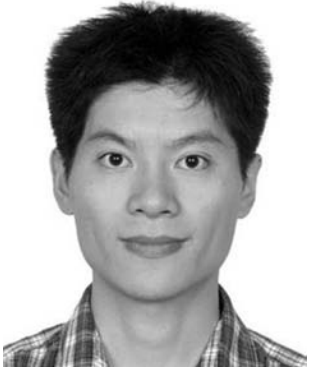
**Hongliang Zhang** was born in Jiangsu, People's Republic of China, in 1981. He received the B.Sc. and Ph.D. degrees in automatic control from the National University of Defense Technology, Changsha, People's Republic of China, in 2004 and 2010, respectively.

He is now with the National University of Defense Technology as a lecturer. His current research interests include inertial navigation and integrated navigation.

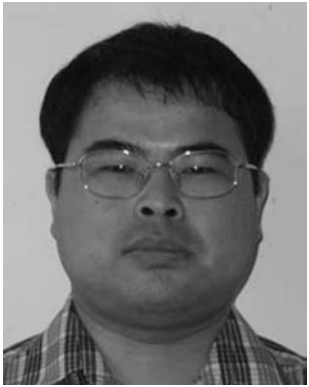
**Meiping Wu** was born in Fujian, People's Republic of China, in 1971. He received the B.Sc. and M.Sc. degrees in aviation mechanics and the Ph.D. degree in navigation, guidance, and control from the Department of Automatic Control, National University of Defense Technology, Changsha, People's Republic of China, in 1993, 1996, and 2000, respectively.

Currently, he is with the National University of Defense Technology as a Professor and Dean of Education of the College of Mechatronics and Automation. He is an expert of the 863 High Technology Project. His scientific interests include aircraft navigation, guidance, and control.

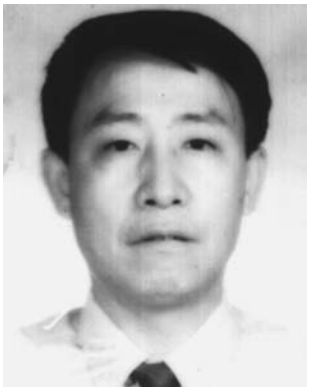
**Xiaoping Hu** was born in Sichuan, People's Republic of China, in 1960. He received the B.Sc. and M.Sc. degrees in automatic control systems and aircraft designing from the Department of Automatic Control, National University of Defense Technology, Changsha, People's Republic of China, in 1982 and 1985, respectively.

Currently, he is with the National University of Defense Technology as a professor and is the Dean of the College of Mechatronics and Automation. His scientific interests include inertial and satellite navigation, aircraft guidance, and control.

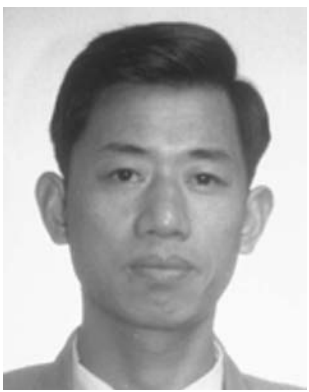
**Dewen Hu** (M'03—SM'06) was born in Hunan, China, in 1963. He received the B.Sc. and M.Sc. degrees from Xi'an Jiaotong University, Xi'an, China, in 1983 and 1986, respectively, and the Ph.D. degree from the National University of Defense Technology, Changsha, China, in 1999, all in automatic control.

Since 1986, he has been at the National University of Defense Technology, where he became a professor in 1996. From October 1995 to October 1996, he was a visiting scholar at the University of Sheffield, United Kingdom. He has authored or coauthored three monographs and more than 200 papers published in international journals and conference proceedings. His research interests include cognitive science, signal processing, system identification and control, and neural networks. He is an action editor of *Neural Networks*.

Dr. Hu is the joint recipient of more than a dozen academic prizes for his research on neuroscience, neurocontrol, and signal processing.